\pdfoutput=1

\documentclass[11pt]{article}

\usepackage[final]{acl}

\usepackage{times}
\usepackage{latexsym}
\usepackage{booktabs}
\usepackage{array}
\usepackage{siunitx} 

\usepackage[T1]{fontenc}

\usepackage[utf8]{inputenc}

\usepackage{microtype}

\usepackage{inconsolata}

\usepackage{graphicx}
\DeclareGraphicsExtensions{.pdf,.png,.jpg}
\usepackage{amsmath}
\usepackage{multicol}
\usepackage{multirow}
\usepackage{amssymb}
\usepackage{tcolorbox}
\usepackage{longtable}

\usepackage{listings}
\usepackage{xcolor} 
\usepackage{subcaption}
\usepackage{float}
\usepackage{placeins}

\title{PrionNER: A Named Entity Recognition Dataset for \\ Prion Disease Biomedical Literature}

\author{
\textbf{An Dao}\textsuperscript{1,5*} \quad
\textbf{Nhan Ly}\textsuperscript{2*} \quad
\textbf{Thao Tran}\textsuperscript{2*} \quad
\textbf{Yuji Matsumoto}\textsuperscript{3,4} \quad
\textbf{Akiko Aizawa}\textsuperscript{5} \\
\textsuperscript{1}The University of Tokyo, Tokyo, Japan \\
\textsuperscript{2}Medical Doctor, Independent Researcher \\
\textsuperscript{3} Center for Language AI Research, Tohoku University, Sendai, Japan \\
\textsuperscript{4} RIKEN Center for Advanced Intelligence Project, Tokyo, Japan \\
\textsuperscript{5} National Institute of Informatics, Tokyo, Japan \\
\texttt{dtan@g.ecc.u-tokyo.ac.jp},
\texttt{trinhanly1996@gmail.com},
\texttt{thaotran1490@gmail.com},\\
\texttt{yuji.matsumoto.a4@tohoku.ac.jp},
\texttt{aizawa@nii.ac.jp}
}
\begin{document}
\maketitle
\renewcommand{\thefootnote}{\fnsymbol{footnote}}
\footnotetext[1]{These authors contributed equally to this work.}
\renewcommand{\thefootnote}{\arabic{footnote}}
\begin{abstract}

Prion diseases are rare, rapidly progressive, and fatal neurodegenerative disorders that remain difficult to diagnose, particularly in their early stages because of nonspecific clinical presentations.
However, to our knowledge, there is no publicly available prion-disease-focused dataset designed to capture a broad range of clinically relevant entities from the biomedical literature.
We introduce \textbf{PrionNER}, a manually annotated named entity recognition dataset for prion disease clinical information in PubMed abstracts.
The current release comprises 317 abstracts, 2,943 sentences, and 6,955 text-bound entity annotations spanning 15 coarse-grained and 31 fine-grained clinically oriented entity types covering diseases, symptoms, diagnostics, findings, anatomy, treatments, and temporal and statistical evidence.
Inter-annotator agreement reaches 81.78 exact-match F1, indicating strong annotation consistency.
We benchmark supervised BERT baselines, W2NER, and zero-shot extractors on PrionNER.
W2NER is the strongest supervised model, and Gemma-4-31B is the strongest zero-shot model, but the benchmark remains challenging, especially for structurally complex mentions and fine-grained clinically adjacent label distinctions.
PrionNER provides a clinically grounded benchmark for prion-disease information extraction and supports research on rare-disease biomedical NLP under low-resource, fine-grained, and non-flat extraction conditions.
The dataset, annotation guidelines, and evaluation scripts are available at \url{https://github.com/daotuanan/PrionNER/}.

\end{abstract}

\section{Introduction}
Prion diseases are rapidly progressive, fatal neurodegenerative disorders caused by the misfolding and accumulation of pathological prion protein~\cite{prusiner1998prions}. 
They remain untreatable while carrying a severe risk of iatrogenic transmission~\cite{cdc2026prions}.
Early diagnosis is difficult because these diseases are rare and therefore unfamiliar to many general physicians, while their initial symptoms are often vague and overlap with other psychiatric and neurological conditions~\cite{geschwind2015prion}.
By the time the disease reaches a clearly recognizable stage, it is often already terminal, limiting opportunities for intervention and increasing the risk of inadvertent spread during routine medical procedures~\cite{vallabh2020treatment}.
Consequently, recent research efforts have emphasized earlier diagnosis both to reduce transmission risk and to identify patient cohorts for future therapeutic trials~\cite{shimamura2025diagnosticcriteria}. 
To achieve this, consolidating scattered clinical knowledge into a unified, foundational dataset is essential.

Named entity recognition (NER) has substantially advanced clinical data extraction by structuring key concepts from biomedical text, including diseases, symptoms, diagnostic tests, and biomarkers~\cite{lee2020biobert,gu2021domain}.
This progress is supported by widely used biomedical and clinical corpora such as NCBI Disease~\cite{dogan2014ncbi}, BC5CDR~\cite{li2016biocreative}, CRAFT~\cite{bada2012concept}, MedMentions~\cite{mohan2019medmentions}, i2b2/VA~\cite{uzuner2011i2b2}, and ShARe/CLEF~\cite{pradhan2013task}.
However, most existing resources target broad biomedical domains, disease mentions, or general classifications rather than the clinically rich diagnostic evidence needed for rare-disease recognition in biomedical literature.
Rare-disease-oriented datasets also focus more narrowly on disease, symptom, sign, or phenotype extraction than on the broader diagnostic schema needed for prion disease~\cite{martinez2022raredis,groza2015automatic,shyr2024rare}.
As a result, there is still no publicly available prion-focused NER benchmark that captures the heterogeneous evidence clinicians rely on when distinguishing prion disease and its subtypes from related conditions.
To address this gap, we introduce \textbf{PrionNER}, a manually annotated dataset for named entity recognition in prion disease clinical narratives derived from PubMed abstracts.

PrionNER contains 317 abstracts with 15 coarse-grained and 31 fine-grained entity types, and its pre-adjudication double-annotated test split reaches 81.78 entity-level exact agreement F1.
We evaluate PrionNER using both supervised biomedical encoders and zero-shot extraction models.
Among supervised models, W2NER is strongest in both coarse-grained and fine-grained \texttt{flat-ner}, reaching 81.86 F1 and 80.46 F1, respectively.
Gemma-4-31B is the strongest zero-shot model in \texttt{flat-ner}, reaching 71.41 coarse-grained F1 and 68.41 fine-grained F1.
However, \texttt{non-flat-ner} remains difficult for all models, with the best F1 scores of W2NER (supervised) reaching 13.48 in coarse evaluation and 13.70 in fine evaluation.
These results show that PrionNER is learnable but still challenging: performance consistently drops from coarse-grained to fine-grained prediction, and the remaining difficulty reflects a long-tailed label distribution, clinically adjacent type distinctions, and nested or discontinuous mentions.

This work provides a foundation for prion-disease information extraction and clinically oriented biomedical NLP.
More broadly, PrionNER illustrates a general challenge in rare-disease biomedical NLP: clinically useful extraction often depends on modeling heterogeneous evidence types, fine-grained diagnostic distinctions, and non-flat mention structures.
In this sense, the dataset is relevant beyond prion disease itself and can serve as a compact case study for building clinically grounded resources in other specialized biomedical subdomains.

\section{Related Work}
\begin{table*}[!t]
\centering
\scriptsize
\setlength{\tabcolsep}{2pt}
\resizebox{\textwidth}{!}{%
\begin{tabular}{>{\raggedright\arraybackslash}p{3.7cm}
                >{\raggedright\arraybackslash}p{1.35cm}
                >{\raggedright\arraybackslash}p{1.6cm}
                >{\raggedright\arraybackslash}p{1.8cm}
                >{\raggedright\arraybackslash}p{3.35cm}
                >{\raggedright\arraybackslash}p{1.9cm}}
\toprule
\textbf{Dataset} & \textbf{Domain} & \textbf{Task} & \textbf{Text Source} & \textbf{Entity Coverage} & \textbf{Size} \\
\midrule
\textsc{JNLPBA}~\cite{collier2004jnlpba} & Biomedical & NER & MEDLINE abstracts & Protein, DNA, RNA, cell line, cell type & 2,404 abstracts \\
\textsc{i2b2/VA 2010}~\cite{uzuner2011i2b2} & Clinical & NER + RE & Patient reports & Problems, tests, treatments, assertions & 871 reports \\
\textsc{CRAFT}~\cite{bada2012concept} & Biomedical & Concept Ann. & Full-text articles & Ontology-linked biomedical concepts & 97 articles \\
\textsc{ShARe/CLEF}~\cite{pradhan2013task} & Clinical & NER + Norm. & Clinical reports & Disorder mentions & 300 reports \\
\textsc{NCBI Disease}~\cite{dogan2014ncbi} & Biomedical & NER + Norm. & PubMed abstracts & Disease mentions linked to MeSH/OMIM & 793 abstracts \\
\textsc{HPO corpora}~\cite{groza2015automatic} & Rare disease & NER + Norm. & PubMed abstracts & Phenotype/HPO concept mentions & 228 abstracts \\
\textsc{BC5CDR}~\cite{li2016biocreative} & Biomedical & NER + RE & PubMed abstracts & Chemicals, diseases, chemical-disease links & 1,500 abstracts \\
\textsc{MedMentions}~\cite{mohan2019medmentions} & Biomedical & Concept Ann. & PubMed abstracts & UMLS-linked biomedical concepts & 4,392 abstracts \\
\textsc{RareDis}~\cite{martinez2022raredis} & Rare disease & NER + RE & Rare-disease texts & Disease, rare disease, symptom, sign, anaphor & 1,041 texts \\
\textbf{PrionNER (ours)} & Prion disease & NER & PubMed abstracts & 15 coarse / 31 fine clinically relevant types & 317 abstracts \\
\bottomrule
\end{tabular}
}
\caption{Comparison of PrionNER with representative prior datasets and related resources.}
\label{tab:related_work_comparison}
\end{table*}
In the biomedical literature, JNLPBA~\cite{collier2004jnlpba}, NCBI Disease Corpus~\cite{dogan2014ncbi}, BC5CDR~\cite{li2016biocreative}, CRAFT~\cite{bada2012concept}, and MedMentions~\cite{mohan2019medmentions} provide benchmark resources for entity recognition, normalization, and concept annotation in MEDLINE or PubMed texts.
In the clinical domain, i2b2/VA~\cite{uzuner2011i2b2}, ShARe/CLEF~\cite{pradhan2013task}, and MedDec~\cite{elgaar2024meddec} provide corpora from patient reports, clinical notes, and discharge summaries that focus on clinically meaningful entities or decision spans such as problems, tests, treatments, disorder mentions, and medical decisions.

Resources closest to our setting are mainly rare-disease or phenotype-oriented.
The HPO corpora~\cite{groza2015automatic} target phenotype concept recognition and normalization, while RareDis~\cite{martinez2022raredis} extends this line with annotations for diseases, rare diseases, symptoms, signs, and anaphoric mentions.
However, based on our preliminary inspection of the released RareDis annotations, prion-disease coverage remains limited, and these datasets do not provide the broader clinically oriented schema needed for prion-disease information extraction.
More broadly, prior resources reflect different tradeoffs in domain breadth and label granularity: corpora such as NCBI Disease and BC5CDR emphasize wider biomedical coverage with a small number of target categories~\cite{dogan2014ncbi,li2016biocreative}, whereas our goal is richer evidence coverage within a single specialized domain.

For prion disease specifically, PDDB~\cite{gehlenborg2009prion} is a structured transcriptomic resource rather than a text annotation benchmark.
To our knowledge, no publicly available dataset currently targets prion-disease named entity recognition in the biomedical literature with a broad clinically oriented annotation schema.
PrionNER fills this gap by providing a manually annotated dataset for prion disease clinical narratives in PubMed abstracts.
We next describe how the corpus was collected, filtered, and annotated.

\section{Dataset Construction}
\subsection{Data Sources}
We constructed the corpus from PubMed abstracts retrieved with a keyword-based Boolean query over the title and abstract fields; the full query is provided in Appendix Section~\ref{app:pubmed_search_query}.
The query combined general and subtype-specific prion disease terms, including \textit{Prion Diseases}, \textit{Creutzfeldt-Jakob Disease}, \textit{CJD}, \textit{sporadic CJD}, \textit{familial/genetic CJD}, \textit{variant CJD}, \textit{iatrogenic CJD}, \textit{Kuru}, \textit{Gerstmann-Straussler-Scheinker}, and \textit{Fatal Familial Insomnia} (\textit{FFI}), with clinically oriented terms such as \textit{diagnosis}, \textit{clinical}, \textit{symptoms}, \textit{case}, \textit{progression}, and \textit{treatment}.
To bias retrieval toward human clinical narratives, we excluded terms commonly associated with animal or basic-science studies, including \textit{mice}, \textit{mouse}, \textit{rat}, \textit{animal}, \textit{cell}, \textit{protein}, and \textit{in vitro}.
The PubMed query returned 3,414 abstracts, and 3,138 remained after basic preprocessing, including removal of records with empty abstracts.

A pilot manual screening of approximately 500 abstracts by two annotators showed that the retrieved set still contained many off-target papers unrelated to clinical prion disease narratives.
Common exclusion cases included basic science, animal or other non-human research, non-clinical analyses such as economic or purely epidemiological studies, and papers in which prion disease was not the main focus.
Based on this pilot review, we defined an operational criterion for \textit{related} abstracts and used GPT-5.4~\cite{openai2026gpt54}, prompted as described in Appendix Section~\ref{app:data_collection_prompting}, to screen the 3,138 preprocessed abstracts and reduce manual review workload.
Under this criterion, abstracts were retained if they primarily concerned human prion disease in a clinical context, including diagnosis, symptoms, disease progression, or treatment.
Across the full 3,138-abstract pool, GPT-5.4 predicted 1,304 abstracts as related and 1,834 as unrelated.
We then manually reviewed abstracts from the screened pool together with the pilot-screened set to confirm relevance and remove duplicates.
In total, we manually reviewed 1,383 abstracts, including 868 abstracts rated as related by GPT-5.4.
Within this set, 772 abstracts were labeled related and 611 were labeled not related by human review.
The GPT-5.4 model achieved 90.60 accuracy and 97.80 recall for the related class; it missed only 17 relevant abstracts but incorrectly marked 113 irrelevant abstracts as related, indicating a high-recall but over-inclusive screening strategy.
This bias was appropriate for corpus construction because missing clinically relevant rare-disease abstracts would have been more costly than forwarding some extra candidates to manual review.
Although we initially intended to annotate all 772 eligible abstracts, practical time constraints limited the current release to 317 annotated abstracts.
We therefore treat the present release as a high-quality first benchmark for this specialized domain rather than an exhaustive survey of all eligible prion-disease abstracts.
Appendix Section~\ref{app:data_collection_prompting} also provides the relevance-filter audit.

\subsection{Entity Schema}
We designed the schema to capture clinically meaningful evidence needed for prion-disease diagnostic reasoning and future knowledge graph construction.
It was developed through pilot annotation and iteratively revised after joint review of 20 shared annotated abstracts by two medical doctors with experience in neurological diseases together with one coordinator, so that the labels remained clinically meaningful while also remaining coherent from an NLP annotation perspective.
Because the dataset is intended to support future clinical decision-support work for prion disease diagnosis, the schema reflects the evidence integration a clinician uses when evaluating a suspected case.
Compared with earlier biomedical and rare-disease corpora, which often focus on narrower mention types such as diseases, chemicals, phenotypes, symptoms, signs, or anaphora, our schema is broader because prion diagnosis depends on combining multiple kinds of diagnostic evidence~\cite{dogan2014ncbi,li2016biocreative,groza2015automatic,martinez2022raredis}.
To reflect this workflow, we organize the schema into three groups: \textit{Case Input}, \textit{Case Diagnosis}, and \textit{Clinical Course and Context}.
\textit{Case Input} captures the information available at a patient's presentation, including \texttt{Age}, \texttt{Symptom}, \texttt{Test\_name}, \texttt{Sequences}, \texttt{Anatomic\_location}, and \texttt{Findings}.
\textit{Case Diagnosis} captures the interpretation of those inputs, including disease names, subtype labels (\texttt{Generic\_Prion}, \texttt{Sporadic\_Prion}, \texttt{Familial\_Prion}, \texttt{Acquired\_Prion}), and alternative conditions considered during evaluation (\texttt{Differential\_Diagnosis}).
\textit{Clinical Course and Context} captures clinically relevant supporting information, including \texttt{Treatment}, \texttt{Complication}, \texttt{Time}, and \texttt{Stats}.
This broad-entity, focused-domain design differs from prior related datasets such as RareDis, which do not combine the same breadth of clinically oriented entities within a single prion-focused annotation schema~\cite{martinez2022raredis}.
We include these categories selectively because of their direct clinical relevance and their potential to support diagnosis in realistic settings.
The schema comprises \textbf{15} coarse-grained types and \textbf{31} fine-grained entity types, and annotation was performed at the mention level using minimal span selection.
Table~\ref{tab:coarse_fine_diagnostic_types} provides a short schema summary for the main text, and Appendix Table~\ref{tab:coarse_fine_diagnostic_types_appendix} provides full definitions with representative examples.

\begin{table*}[t]
\centering
{\fontsize{7.5}{8.5}\selectfont
\setlength{\tabcolsep}{3pt}
\begin{tabular}{>{\raggedright\arraybackslash}m{2.4cm}
                >{\raggedright\arraybackslash}m{4.8cm}
                >{\raggedright\arraybackslash}m{7.8cm}}
\specialrule{1pt}{0pt}{0pt}
\textbf{Coarse-grained Type} & \textbf{Fine-grained Types} & \textbf{Definition Summary} \\
\specialrule{1pt}{0pt}{0pt}
\multicolumn{3}{l}{\textbf{Case Input~\cite{willis2006medical}}} \\
\hline
Age & Age & Age or age-at-onset expressions. \\
\hline
Symptom & Symptom & Symptoms, signs, and other explicit clinical manifestations. \\
\hline
Test\_name & Imaging\_test, Electrophysio\_test, Blood\_biomarker\_test, Genetic\_test, Molecular\_assay, Autopsy & Diagnostic procedures, assays, and examination types. \\
\hline
Sequences & Imaging\_sequence & Imaging acquisition or sequence terms. \\
\hline
Anatomic\_location & Anatomic\_location & Anatomical sites linked to findings or symptoms. \\
\hline
Findings & Imaging\_finding, Autopsy\_finding & Radiologic and pathological findings. \\
\hline
\multicolumn{3}{l}{\textbf{Case Diagnosis~\cite{wadoh2022prions}}} \\
\hline
Generic\_Prion & Generic\_Prion & Prion disease mentions without explicit subtype specification. \\
\hline
Sporadic\_Prion & sCJD, sFI, VPSPr & Sporadic prion disease subtypes. \\
\hline
Familial\_Prion & fCJD, GSS, FFI & Familial or inherited prion disease subtypes. \\
\hline
Acquired\_Prion & vCJD, iCJD, Kuru & Acquired prion disease subtypes. \\
\hline
Differential\_Diagnosis & Differential\_Diagnosis & Non-prion diseases considered in differential diagnosis. \\
\hline
\multicolumn{3}{l}{\textbf{Clinical Course and Context~\cite{thabane2013tutorial,willis2006medical,murad2023association}}} \\
\hline
Treatment & Treatment & Therapies, medications, and care strategies. \\
\hline
Complication & Complication & Secondary conditions or end-stage outcomes. \\
\hline
Time & Duration, Time\_point & Temporal spans and specific time anchors. \\
\hline
Stats & Sensitivity, Specificity, Prevalence, Incidence & Diagnostic and epidemiological quantitative measures. \\
\specialrule{1pt}{0pt}{0pt}
\end{tabular}
\caption[Short summary of the PrionNER entity schema.]{Short summary of the PrionNER entity schema.
Full type definitions and representative examples are provided in Appendix Table~\ref{tab:coarse_fine_diagnostic_types_appendix}.}
\label{tab:coarse_fine_diagnostic_types}
\par}
\end{table*}

\subsection{Annotation Workflow and Quality Control}
\label{sec:annotation}

\textbf{Pilot annotation.} Annotation began with a pilot phase of 20 PubMed abstracts.
During this stage, two annotators independently labeled entity mentions in order to identify ambiguities in span boundaries, type assignment, and difficult clinical expressions.
Disagreements were reviewed jointly, and these pilot annotations were used to establish the initial annotation guidelines before large-scale annotation.
To preserve anonymity, we refer to the two annotators as \textit{Annotator 1} and \textit{Annotator 2}.

\textbf{Training annotation and guideline refinement.}
Using the pilot guidelines, the remaining training abstracts were annotated individually, with 151 abstracts annotated by \textit{Annotator 1} and 96 abstracts annotated by \textit{Annotator 2}.
During this stage, the guidelines were iteratively refined to address recurring ambiguities and improve consistency across the corpus.

\textbf{Double annotation, agreement measurement, and final test-set selection.}
After the training-stage guideline refinement, an additional 70 abstracts were independently annotated by both annotators.
Inter-annotator agreement was computed on these independent annotations before any disagreement resolution.
Only after agreement measurement did the annotators jointly adjudicate disagreements in this 70-abstract subset to produce a single finalized annotation layer containing 787 sentences and 1,806 text-bound entity annotations.
We designated this double-annotated and adjudicated subset as the test split because it is the highest-confidence evaluation subset in the corpus: both annotators labeled these abstracts independently, agreement was measured on the pre-adjudication annotations, and disagreements were then resolved to create the final released test annotations.
The clarified adjudication decisions were then propagated back to the training annotations to ensure consistency with the final guideline version, and the final dataset was validated for annotation integrity, including span correctness, label consistency, and the absence of duplicate or conflicting entity annotations.


\section{Dataset Statistics}

\subsection{Corpus Overview}

Overall, PrionNER contains 317 abstracts, 2,943 sentences, and 6,955 text-bound entity annotations across the training and test splits.
The dataset covers 15 coarse-grained and 31 fine-grained entity types spanning diagnostically relevant clinical evidence in prion disease literature.
The training split contains 247 abstracts, and the finalized test split contains 70 abstracts.
As described in Section~\ref{sec:annotation}, the released test split reflects a single merged annotation layer derived from the adjudicated 70-abstract double-annotated subset.
To support reproducibility, Appendix Section~\ref{app:data_collection_prompting}, Appendix Section~\ref{app:annotation_guidelines}, and Appendix Sections~\ref{app:supervised_model_details}--\ref{app:unsupervised_prompting} document the PubMed search query, the abstract-screening prompt, the annotation guidelines, and the model settings used in this work.
The dataset annotations, annotation guidelines, and evaluation scripts are publicly available at \url{https://github.com/daotuanan/PrionNER/}.

\begin{table}[H]
\centering
\footnotesize
\setlength{\tabcolsep}{5pt}
\begin{tabular}{lrrr}
\toprule
\textbf{Split} & \textbf{Abstracts} & \textbf{Sentences} & \textbf{Entities} \\
\midrule
Train & 247 & 2,156 & 5,149 \\
Test  & 70 & 787 & 1,806 \\
\midrule
Total & 317 & 2,943 & 6,955 \\
\bottomrule
\end{tabular}
\caption{Corpus-level statistics of PrionNER.}
\label{tab:dataset_statistics}
\end{table}

\subsection{Inter-Annotator Agreement}

\textbf{Inter-annotator agreement.}
We report agreement on the pre-adjudication independent annotations of the 70-abstract double-annotated subset described in Section~\ref{sec:annotation}.
Agreement is assessed using three complementary symmetric measures: entity-level exact agreement F1, Jaccard similarity, and Cohen's kappa.
Entity-level exact agreement F1 summarizes span-level agreement under the standard NER criterion that an entity counts as correct only when both annotators select the same text span and assign the same entity type~\cite{tjongkimsang2003conll}.
Jaccard similarity measures the overlap between the two annotators' entity sets as the size of the intersection relative to the union~\cite{jaccard1901etude}.
Cohen's kappa is computed on token-level BIO labels to quantify agreement while correcting for agreement expected by chance~\cite{cohen1960kappa}.
On the 70 compared documents, entity-level exact agreement is 81.78 F1.
Jaccard similarity over the annotated entity sets is 69.17, and token-level BIO Cohen's kappa is 81.05.
Together, these metrics provide complementary views of annotation consistency at both the span and token levels, and Appendix Table~\ref{tab:per_label_agreement} provides the full per-label agreement breakdown.

\subsection{Label Distribution and Structural Complexity}

\begin{table}[t]
\centering
\small
\resizebox{\columnwidth}{!}{%
\begin{tabular}{>{\raggedright\arraybackslash}p{0.34\linewidth}rrrr}
\toprule
\textbf{Entity Type} & \textbf{Train n} & \textbf{Train \%} & \textbf{Test n} & \textbf{Test \%} \\
\midrule
\multicolumn{5}{l}{\textit{Most Frequent}} \\
Symptom & 791 & 16.99 & 248 & 15.03 \\
Generic\_Prion & 741 & 15.92 & 233 & 14.12 \\
Anatomic\_location & 646 & 13.88 & 223 & 13.52 \\
Imaging\_test & 261 & 5.61 & 75 & 4.55 \\
Duration & 217 & 4.66 & 49 & 2.97 \\
\midrule
\multicolumn{5}{l}{\textit{Least Frequent Observed}} \\
\textit{sFI} & 3 & 0.06 & 0 & 0.00 \\
Prevalence & 4 & 0.09 & 1 & 0.06 \\
Incidence & 6 & 0.13 & 4 & 0.24 \\
Sensitivity & 0 & 0.00 & 4 & 0.24 \\
Specificity & 0 & 0.00 & 4 & 0.24 \\
\bottomrule
\end{tabular}
}
\caption[Selected most frequent and least frequent observed schema-defined fine-grained entity types in the train and test splits.]{Selected most frequent and least frequent observed schema-defined fine-grained entity types in the train and test splits.
The full split-wise fine-grained distribution is provided in Appendix Table~\ref{tab:checked_finegrained_distribution_train_test}.}
\label{tab:top_entity_types}
\end{table}

\begin{figure}[t]
\centering
\includegraphics[width=0.99\linewidth]{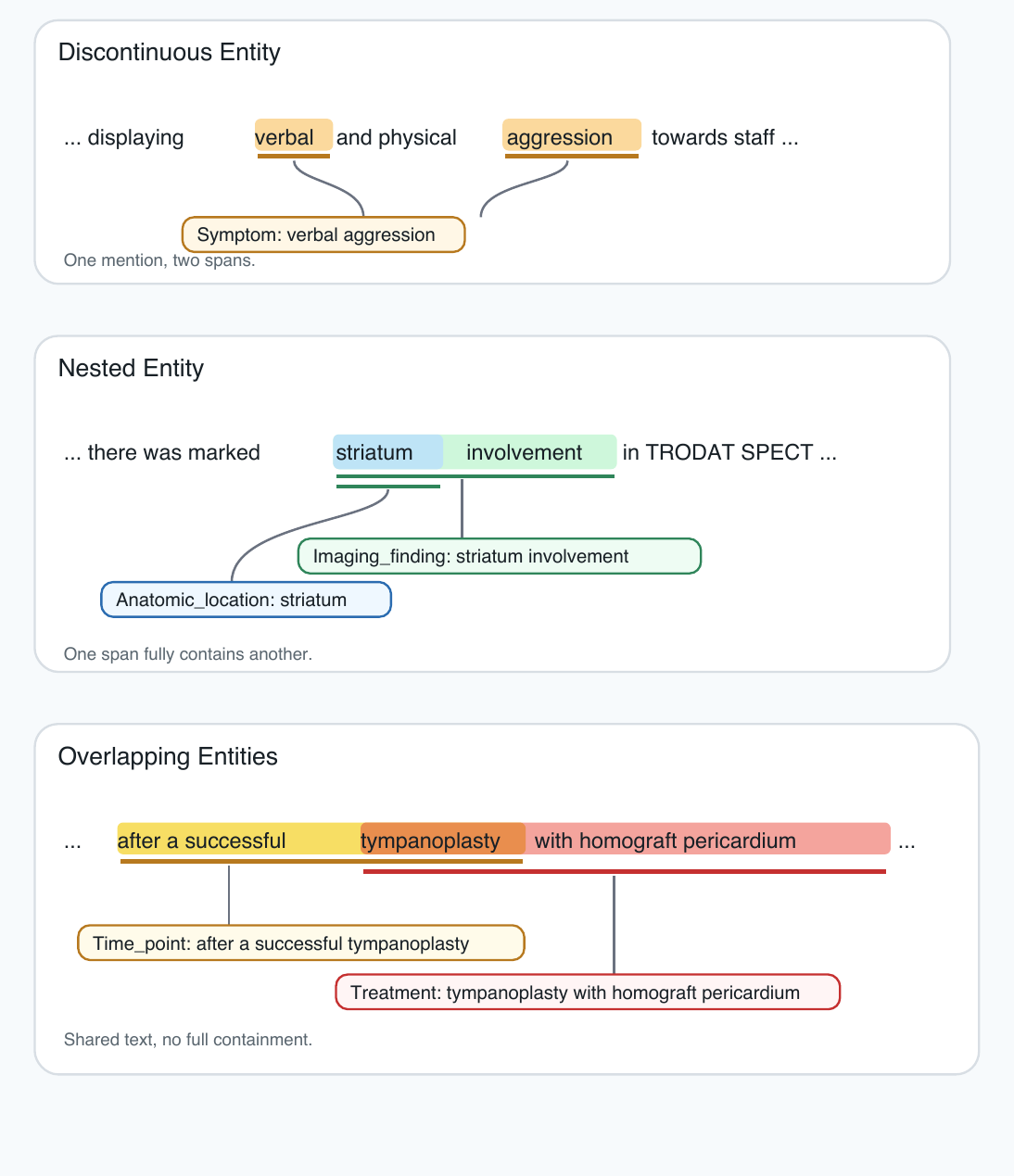}
\caption{Representative discontinuous, nested, and overlapping entity annotations from the test set.}
\label{fig:special_entity_examples}
\end{figure}

Tables~\ref{tab:dataset_statistics} and~\ref{tab:top_entity_types} show that PrionNER covers a broad range of diagnostically relevant evidence while remaining strongly long-tailed.
The most frequent fine-grained labels are \textit{Symptom}, \textit{Generic\_Prion}, and \textit{Anatomic\_location}, whereas several clinically meaningful types remain rare.
The adjudicated test set also retains structurally complex mentions, including 34 discontinuous entities, 80 nested pairs, and 2 overlapping pairs.
Figure~\ref{fig:special_entity_examples} gives representative examples of these discontinuous, nested, and overlapping annotations from the test set.
Full split-wise label counts and structural statistics are provided in Appendix Section~\ref{app:additional_dataset_statistics}.
With the corpus characteristics established, we next describe the experimental setup used to benchmark the dataset.

%
%
\section{Experiments}
\subsection{Experimental Settings}
We formulate PrionNER as a sequence-labeling task over schema-defined entity mentions.
For supervised models, this sequence-labeling setup is implemented with standard BIO tagging, where each entity type is paired with \texttt{B-} and \texttt{I-} prefixes together with the outside label \texttt{O} (e.g., \texttt{B-Symptom}, \texttt{I-Symptom}, \texttt{O}).
Models are trained on the training split and evaluated on the test set, with corpus statistics summarized in Table~\ref{tab:dataset_statistics}.
Because the corpus is relatively small and highly long-tailed, we do not introduce a separate development split, as doing so would further reduce training coverage for already sparse labels.
Instead, we use fixed model configurations applied uniformly within each model family, and no hyperparameters are selected based on test-set performance.
We evaluate the corpus under two structural settings.
In the \texttt{flat-ner} setting, both training and evaluation are restricted to the flat-compatible portion of the corpus, so only contiguous, non-overlapping gold entities are included.
In the \texttt{non-flat-ner} setting, evaluation is restricted to the structurally complex portion of the corpus, so the gold annotations include only nested, discontinuous, and overlapping entities.
We evaluate the dataset under two label-granularity settings.
In the \textit{fine-grained} setting, models are trained and evaluated using the 31 schema-defined entity types.
In the \textit{coarse-grained} setting, models are trained and evaluated separately using the 15 coarse-grained categories.
This coarse setting is implemented as a separate task rather than by collapsing fine-grained predictions at evaluation time.
We report entity-level precision, recall, and F1-score using exact span-and-label matching.
Our primary metric is micro-averaged F1 on the test set, and we report it for both coarse-grained and fine-grained settings.
The search query and screening prompt are documented in Appendix Section~\ref{app:data_collection_prompting}, and model configurations are documented in Appendix Sections~\ref{app:supervised_model_details} and~\ref{app:unsupervised_prompting}, so that the dataset construction and evaluation pipeline can be reproduced more directly.

\subsection{Models}

\subsubsection{Supervised Models}

We evaluate supervised baselines fine-tuned on the PrionNER training data.
These include BioBERT~\cite{lee2020biobert}, ClinicalBERT~\cite{alsentzer2019publicly}, and PubMedBERT~\cite{gu2021domain}, which are widely used pretrained encoders for biomedical and clinical text mining, together with W2NER~\cite{li2022unified} as an additional structured supervised baseline.
We include W2NER because it is a supervised model that can also handle the \texttt{non-flat-ner} setting, allowing a supervised comparison beyond standard BIO-style flat tagging.
For W2NER, we use PubMedBERT embeddings as the underlying biomedical text representation.
The encoder baselines are implemented as BIO-based token classification systems and trained under the same data and model-selection protocol.
The BERT baselines operate only in the standard \texttt{flat-ner} setting, whereas W2NER is additionally evaluated in the \texttt{non-flat-ner} setting.
Detailed checkpoints and training hyperparameters are described in Appendix Section~\ref{app:supervised_model_details}.

\subsubsection{Zero-shot Models}

We additionally evaluate extraction performance using model families that do not require task-specific supervised fine-tuning on PrionNER.
These models include OpenAI GPT-5.4~\cite{openai2026gpt54}, Gemma 4 models~\cite{team2024gemma}, the GLiNER2 variants GLiNER2-short and GLiNER2-def~\cite{zaratiana2025gliner2}, and GLiNER-BioMed~\cite{yazdani2025gliner}.
Because these systems typically return entity strings rather than reliable character offsets, their outputs require a deterministic span-alignment step before exact-match evaluation.
For GLiNER2, we consider two input modes: GLiNER2-short, which provides only entity type names, and GLiNER2-def, which provides entity type names together with their definitions.
GLiNER-BioMed is evaluated only in the \textit{short} setting.
In our current GLiNER-based pipeline, non-flat behavior can be represented for nested and overlapping entities through multiple contiguous spans, but the model does not support explicit discontinuous multi-span entity objects.
We also explored additional small local biomedical LLMs, including \texttt{Llama3-OpenBioLLM-8B} and \texttt{BioMistral-7B}, but found them unreliable for strict exact-span extraction in this pipeline.
This setting allows us to assess how well general or biomedical zero-shot extractors transfer to rare-disease clinical literature without supervised adaptation.
Detailed model checkpoints, prompting, and inference settings are described in Appendix Section~\ref{app:unsupervised_prompting}.

\subsection{Main Results}

Tables~\ref{tab:main_results_flat} and~\ref{tab:main_results_nested_discontinuous} report the main results under coarse-grained and fine-grained label spaces in the \texttt{flat-ner} and \texttt{non-flat-ner} settings.
All models are evaluated in \texttt{flat-ner}, whereas \texttt{non-flat-ner} is reported only for W2NER and the zero-shot models because the BERT baselines operate only in the standard flat setting.
This design keeps discontinuous and other structurally complex gold annotations in the benchmark rather than discarding them.

\begin{table}[t]
\centering
{\fontsize{5.8}{6.6}\selectfont
\setlength{\tabcolsep}{2.5pt}
\renewcommand{\arraystretch}{0.92}
\resizebox{\columnwidth}{!}{%
\begin{tabular}{lrrr|rrr}
\toprule
\multirow{2}{*}{\textbf{Model}} & \multicolumn{3}{c|}{\textbf{Coarse}} & \multicolumn{3}{c}{\textbf{Fine}} \\
\cmidrule(lr){2-4} \cmidrule(lr){5-7}
 & \textbf{P} & \textbf{R} & \textbf{F1} & \textbf{P} & \textbf{R} & \textbf{F1} \\
\midrule
\multicolumn{7}{l}{\textit{Supervised}} \\
\midrule
W2NER & \textbf{84.23} & 79.63 & \textbf{81.86} & \textbf{83.07} & 78.01 & \textbf{80.46} \\
PubMedBERT & 81.88 & \textbf{81.37} & 81.62 & 78.93 & \textbf{80.06} & 79.49 \\
BioBERT & 80.96 & 81.47 & 81.22 & 78.61 & 78.61 & 78.61 \\
ClinicalBERT & 79.45 & 79.10 & 79.28 & 78.29 & 78.73 & 78.51 \\
\midrule
\multicolumn{7}{l}{\textit{Zero-shot}} \\
\midrule
GPT-5.4 & 72.50 & 58.81 & 64.94 & 72.01 & 58.42 & 64.50 \\
Gemma-4-31B & \textbf{78.33} & \textbf{65.61} & \textbf{71.41} & \textbf{80.14} & \textbf{59.67} & \textbf{68.41} \\
Gemma-4-26B & 77.43 & 55.25 & 64.48 & 79.71 & 46.67 & 58.87 \\
GLiNER2-def & 67.62 & 59.41 & 63.25 & 73.29 & 43.10 & 54.28 \\
GLiNER-BioMed & 29.50 & 42.71 & 34.90 & 28.99 & 45.35 & 35.37 \\
GLiNER2-short & 32.63 & 32.67 & 32.65 & 39.51 & 34.79 & 37.00 \\
\bottomrule
\end{tabular}
}
\caption{\texttt{flat-ner} results for all supervised and zero-shot models on PrionNER under coarse-grained and fine-grained label spaces. Best scores are bolded separately within the supervised and zero-shot sections.}
\label{tab:main_results_flat}
\renewcommand{\arraystretch}{1}
}
\end{table}

\begin{table}[t]
\centering
{\fontsize{5.8}{6.6}\selectfont
\setlength{\tabcolsep}{2.5pt}
\renewcommand{\arraystretch}{0.92}
\resizebox{\columnwidth}{!}{%
\begin{tabular}{lrrr|rrr}
\toprule
\multirow{2}{*}{\textbf{Model}} & \multicolumn{3}{c|}{\textbf{Coarse}} & \multicolumn{3}{c}{\textbf{Fine}} \\
\cmidrule(lr){2-4} \cmidrule(lr){5-7}
 & \textbf{P} & \textbf{R} & \textbf{F1} & \textbf{P} & \textbf{R} & \textbf{F1} \\
\midrule
\multicolumn{7}{l}{\textit{Supervised / Structured}} \\
\midrule
W2NER & \textbf{12.40} & \textbf{14.81} & \textbf{13.48} & \textbf{12.74} & \textbf{14.81} & \textbf{13.70} \\
\midrule
\multicolumn{7}{l}{\textit{Zero-shot}} \\
\midrule
GPT-5.4 & 4.19 & 9.63 & 5.84 & 4.19 & \textbf{9.63} & 5.84 \\
Gemma-4-31B & 4.82 & \textbf{12.59} & 6.97 & \textbf{4.96} & \textbf{9.63} & \textbf{6.55} \\
Gemma-4-26B & \textbf{5.76} & 11.85 & \textbf{7.75} & 0.56 & 0.74 & 0.63 \\
GLiNER2-def & 3.53 & 4.44 & 3.93 & 2.78 & 0.74 & 1.17 \\
GLiNER-BioMed & 0.00 & 0.00 & 0.00 & 0.00 & 0.00 & 0.00 \\
GLiNER2-short & 2.07 & 2.22 & 2.14 & 2.90 & 2.96 & 2.93 \\
\bottomrule
\end{tabular}
}
\caption{\texttt{non-flat-ner} results on PrionNER under coarse-grained and fine-grained label spaces. This setting includes only nested, discontinuous, and overlapping entities. Best scores are bolded separately within the supervised/structured and zero-shot sections.}
\label{tab:main_results_nested_discontinuous}
\renewcommand{\arraystretch}{1}
}
\end{table}

In the \texttt{flat-ner} setting, W2NER is the strongest supervised model in both label spaces, reaching 81.86 F1 in the coarse-grained setting and 80.46 F1 in the fine-grained setting.
Among the zero-shot systems, Gemma-4-31B performs best, with 71.41 coarse-grained F1 and 68.41 fine-grained F1.

A notable pattern in \texttt{flat-ner} is the relatively high precision of the strongest zero-shot models under the span-recovery pipeline described in Appendix Section~\ref{app:unsupervised_prompting}.
Gemma-4-31B reaches 78.33 precision in the coarse-grained setting and 80.14 precision in the fine-grained setting, suggesting that the schema is semantically coherent enough for strong zero-shot label assignment from definitions and task instructions alone, even though recall remains well below the supervised baselines.
Appendix Sections~\ref{app:unsupervised_prompting} and~\ref{app:entity_only_confusion_analysis} further suggest that the strongest zero-shot model is limited more by omission than by label precision: Gemma-4-31B remains close to PubMedBERT in entity-only precision while trailing much more clearly in recall.

In the \texttt{non-flat-ner} setting, performance drops sharply for all models, but W2NER remains clearly stronger than the zero-shot systems, reaching 13.48 coarse-grained F1 with 12.40 precision and 14.81 recall, and 13.70 fine-grained F1 with 12.74 precision and 14.81 recall.
Among the zero-shot systems, the best coarse-grained \texttt{non-flat-ner} score is 7.75 for Gemma-4-26B, while the best fine-grained score is 6.55 for Gemma-4-31B.
GPT-5.4 is second among these models in the fine-grained \texttt{non-flat-ner} setting with 5.84 F1 and remains competitive in the coarse-grained setting, but it does not lead there.
Within the GLiNER family, coarse-grained \texttt{non-flat-ner} performance improves slightly relative to fine-grained evaluation, but the GLiNER variants remain clearly behind the top LLMs, and GLiNER-BioMed fails to recover any non-flat entities in this setting.

Overall, fine-grained prediction is generally harder than coarse-grained prediction, confirming that finer label distinctions remain substantially harder, especially once structurally complex mentions are retained in evaluation.
Appendix Figure~\ref{fig:fine_precision_recall_scatter} further visualizes the fine-grained precision--recall trade-offs across the evaluated models.

\subsection{Per-Type Analysis}

Appendix Figure~\ref{fig:fine_per_entity_f1_heatmap} summarizes the per-type fine-grained F1 patterns for the main supervised and zero-shot models, and Appendix Table~\ref{tab:per_type_results_main} gives the full scores.
Performance is uneven across the schema.
Supervised models dominate many frequent labels such as \textit{Anatomic\_location}, \textit{Generic\_Prion}, \textit{Imaging\_test}, \textit{Symptom}, and \textit{sCJD}, which helps explain their strong aggregate performance.
Zero-shot models remain competitive on several semantically clearer or clinically distinctive labels, with GPT-5.4 or Gemma 4 achieving the best score for types such as \textit{FFI}, \textit{Electrophysio\_test}, \textit{fCJD}, \textit{Genetic\_test}, \textit{Molecular\_assay}, and \textit{Differential\_Diagnosis}.
GLiNER2-def is a notable exception within the GLiNER family, reaching 1.00 F1 on \textit{Sensitivity}, but sparse or boundary-sensitive labels such as \textit{Time\_point}, \textit{iCJD}, \textit{Imaging\_finding}, and especially \textit{Prevalence} remain difficult for nearly all systems.
These extreme per-type values should be interpreted cautiously for very rare labels, where isolated perfect or zero scores may reflect only a handful of test mentions rather than stable schema-wide behavior.

\section{Discussions}
The main results and per-type patterns motivate a broader discussion of why the benchmark remains difficult and how well it aligns with prion-disease clinical evidence.

\subsection{Remaining Challenges}

First, the entity distribution is long-tailed.
As shown in Table~\ref{tab:top_entity_types}, a small number of frequent labels account for a large share of mentions: \textit{Symptom}, \textit{Generic\_Prion}, and \textit{Anatomic\_location} account for 791, 741, and 646 mentions in the training split and 248, 233, and 223 in the test split, respectively.
At the same time, many clinically meaningful categories remain sparse, including \textit{Prevalence} (4 in the training split, 1 in the test split) and \textit{Sensitivity} and \textit{Specificity} (0 in the training split and 4 each in the test split).
This skew is not merely a corpus artifact but reflects the structure of real diagnostic reporting, where a few common evidence types dominate while rarer findings and subtypes remain clinically important.
Difficulty also reflects lexical diversity rather than frequency alone: even common labels such as \textit{Symptom}, \textit{Anatomic\_location}, and \textit{Duration} cover hundreds of distinct normalized surface forms in the training split.
There is also substantial train--test surface-form mismatch, so models must generalize beyond memorized mention dictionaries even for relatively common categories.

Second, the schema requires distinctions among diagnostically adjacent categories, including prion subtypes, test names, findings, temporal expressions, and differential diagnoses, so errors are not only boundary errors but often clinically meaningful type confusions.
In practice, subtype mentions such as \textit{fCJD}, \textit{iCJD}, and ``corneal transplant-related CJD'' are sometimes predicted as \textit{Generic\_Prion}.
Others arise when late-stage downstream consequences such as \textit{dysphagia} or \textit{death} appear lexically symptom-like even though they are better labeled as \textit{Complication} in context.
The imaging cluster is also challenging, with \textit{Imaging\_test}, \textit{Imaging\_sequence}, and \textit{Imaging\_finding} often requiring fine-grained distinction among modality names, acquisition terms, and radiologic abnormalities.
Appendix Section~\ref{app:entity_only_confusion_analysis} provides a more detailed entity-only confusion analysis with illustrative examples.

Third, the corpus contains discontinuous, nested, and overlapping mentions that are difficult for standard extraction pipelines.
This difficulty is partly a consequence of dataset design itself: PrionNER preserves clinically adjacent labels and non-flat mention structures because simplifying them away would reduce the realism of the target task.
At the structural level, the training split contains 97 discontinuous entities, 235 nested pairs, and 5 overlapping pairs, while the test split contains 34 discontinuous entities, 80 nested pairs, and 2 overlapping pairs.
In practice, this means that most structural difficulty comes from nesting rather than true overlap.
Appendix Tables~\ref{tab:test_structural_statistics}--\ref{tab:overlapping_entity_statistics} provide the full structural breakdown, and Figure~\ref{fig:special_entity_examples} illustrates representative examples from the test set.

\subsection{Clinical Alignment}

PrionNER captures key prion-disease characteristics, including symptoms, tests, and disease progression described in the academic literature and clinical guideline documents, rather than arbitrary lexical patterns~\cite{wadoh2022prions,cdc2026prions}.
CJD-related terminology is strongly represented in the labeled terms, with \textit{CJD} appearing 576 times, which is consistent with the prominence of Creutzfeldt-Jakob disease in scientific reports.
The \texttt{Symptom} entity (1,279 mentions) aligns closely with current medical guidance and captures hallmark clinical features such as \textit{dementia} (79), \textit{myoclonus} (79), and \textit{ataxia} (47).
The dataset also preserves rarer but clinically important prion-disease subtypes and variants such as FFI (85), GSS (63), and the Heidenhain variant (4).
Finally, the \texttt{Duration} entity (290 mentions) captures meaningful disease-course information, with frequent expressions such as \textit{within a year} (7) and \textit{12 months} (7), reflecting the typical one-year survival duration described in the clinical literature.
Taken together, these patterns show that PrionNER aligns well with expert knowledge and published guidelines.
This makes the dataset useful not only for benchmarking NER systems but also for downstream knowledge extraction and future diagnostic-support settings.

\section{Conclusions}
In this paper, we introduce PrionNER, a manually annotated named entity recognition dataset for prion disease biomedical literature derived from 317 PubMed abstracts.
We present a clinically grounded schema that captures fine-grained diagnostic evidence and non-flat entity structure, together with a benchmark spanning supervised and zero-shot extraction settings.
PrionNER provides a reliable annotation resource, with 81.78 entity-level exact agreement F1 on the pre-adjudication double-annotated test split.
Our experiments show that W2NER provides the strongest supervised results, PubMedBERT is the strongest BERT baseline, and Gemma-4-31B is the strongest zero-shot model, while the remaining performance gaps confirm that PrionNER is a useful but challenging benchmark for prion-disease information extraction.
More broadly, the benchmark surfaces challenges that are likely to recur in other rare-disease settings, including sparse but clinically important labels, fine-grained diagnostic distinctions, and nested or discontinuous spans.
We hope PrionNER will support future research on rare-disease information extraction, structured knowledge construction, and clinically oriented biomedical NLP, including relation extraction, document-level evidence consolidation, and normalization or retrieval settings that connect extracted mentions to structured rare-disease knowledge resources.

\section*{Limitations}
PrionNER has several limitations.
First, it is built from 317 PubMed abstracts, which limits corpus size and coverage.
Second, its focus on prion diseases may reduce generalizability to other disorders.
Third, the data come from biomedical abstracts rather than clinical notes, so the language is more summarized and structured than real-world records.
Future work should expand the corpus, improve coverage of sparse labels, and incorporate additional clinical text sources.

\section*{Ethics Statement}
This study uses publicly available PubMed abstracts and does not involve human subjects or protected health information.
No identifiable personal data were collected or processed, so institutional review board approval was not required.

\section*{Acknowledgement}
This work was supported by Cross-ministerial Strategic Innovation Promotion Program (SIP) on “Integrated Health Care System” Grant Number JPJ012425.

\bibliography{custom}

\clearpage
\appendix
\section{Data Collection and Model Setup}
\label{app:data_collection_prompting}

\subsection{PubMed Search Query}
\label{app:pubmed_search_query}

The following Boolean query is used to retrieve candidate abstracts from PubMed:

\noindent
\begin{minipage}{\columnwidth}
\begin{lstlisting}[basicstyle=\ttfamily\scriptsize,breaklines=true,breakatwhitespace=false,columns=fullflexible,keepspaces=true]
(
  "Prion Diseases"[Title/Abstract] OR
  "Creutzfeldt-Jakob Disease"[Title/Abstract] OR
  "CJD"[Title/Abstract] OR
  "sporadic CJD"[Title/Abstract] OR
  "familial CJD"[Title/Abstract] OR
  "genetic CJD"[Title/Abstract] OR
  "variant CJD"[Title/Abstract] OR
  "iatrogenic CJD"[Title/Abstract] OR
  "Kuru"[Title/Abstract] OR
  "Gerstmann-Straussler-Scheinker"[Title/Abstract] OR
  "Fatal Familial Insomnia"[Title/Abstract] OR
  "FFI"[Title/Abstract]
)
AND
(
  diagnosis[Title/Abstract] OR
  clinical[Title/Abstract] OR
  symptoms[Title/Abstract] OR
  case[Title/Abstract] OR
  progression[Title/Abstract] OR
  treatment[Title/Abstract]
)
NOT
(
  mice[Title/Abstract] OR
  mouse[Title/Abstract] OR
  rat[Title/Abstract] OR
  animal[Title/Abstract] OR
  cell[Title/Abstract] OR
  protein[Title/Abstract] OR
  in vitro[Title/Abstract]
)
\end{lstlisting}
\end{minipage}

\subsection{Prompt Used for Abstract Relevance Screening}

Before applying model-based screening to the full retrieval set, two annotators manually screened approximately 500 abstracts and observed that a large fraction of the keyword-matched results were not actually suitable for corpus construction.
The most common false matches were papers centered on basic science, animal or other non-human studies, non-clinical analyses such as economic or purely epidemiological reports, or papers in which prion disease was only a secondary topic.
We used this pilot review to formulate an operational definition of \textit{related} abstracts for automated screening.

The GPT-5.4 screening prompt asks the model to determine whether an abstract is relevant to \textit{human prion diseases in a clinical context}, based only on the abstract text.
Relevant abstracts are defined as those primarily focused on diagnosis, symptoms, disease progression, or treatment in human prion disease.
The prompt explicitly instructed the model to reject abstracts centered on basic science, animal or non-human studies, protein mechanisms without clinical human focus, unrelated primary diseases, or non-clinical reports such as economic, policy, surveillance, or purely epidemiological analyses.

The model is required to return a strict JSON object with three fields: `is\_relevant`, `reason`, and `evidence\_spans`.
If relevance checking is enabled, we verify that the response is a JSON object with \texttt{is\_relevant} as a Boolean value, \texttt{reason} as a non-empty string, and \texttt{evidence\_spans} as a list.
We then retain only evidence spans that can be matched to the source abstract after light text normalization.
A simplified version of the prompt structure is shown below.

\begin{lstlisting}[basicstyle=\ttfamily\footnotesize,breaklines=true,breakatwhitespace=false,columns=fullflexible,keepspaces=true]
System: You are a biomedical abstract screening assistant.
Decide whether the abstract is relevant to human prion
diseases in a clinical context.
Return one strict JSON object only.

User: 
- Use the provided relevance definition. 
- Judge based only on the abstract text. 
- Exclude basic science, animal, non-human, unrelated-topic, and non-clinical abstracts. 
- Return JSON only in the form:


{
  "is_relevant": true,
  "reason": "<short reason>",
  "evidence_spans": [
    "<short exact span from abstract>"
  ]
}
\end{lstlisting}

\subsection{Audit of the Relevance Filter}

After the initial retrieval and preprocessing stage produced 3,138 abstracts, GPT-5.4 screened the full set with the relevance definition above and labeled 1,304 abstracts as related and 1,834 as unrelated.
We then manually reviewed abstracts from the screened pool together with the earlier pilot screening results to confirm relevance and remove duplicates.
In total, this yielded 1,383 manually reviewed abstracts for auditing the screening process, including 868 abstracts rated as related by GPT-5.4.

Within this manually reviewed set, 772 abstracts were judged truly related and 611 were judged truly unrelated.
GPT-5.4 predicted 868 abstracts as related and 515 as unrelated.
Among the 868 abstracts rated as related by GPT-5.4, 755 were truly related and 113 were actually unrelated, corresponding to a precision of 86.98 for the related class.
The resulting confusion pattern was strongly asymmetric: only 17 truly related abstracts were missed, whereas 113 unrelated abstracts were incorrectly marked as related.
This shows that GPT-5.4 was aggressive in calling abstracts relevant.

On this 1,383-abstract audit set, GPT-5.4 achieved 90.60 accuracy and 89.65 balanced accuracy.
For the related class, precision was 86.98, recall was 97.80, and F1 was 92.07.
For the unrelated class, precision was 96.70, recall was 81.51, and F1 was 88.45.
These headline metrics indicate that about 90.60\% of all abstracts were classified correctly.
Balanced accuracy of 89.65\% is a useful summary here because it averages recall on the related and unrelated classes, which is important given the model's asymmetric behavior.
For the related class, GPT-5.4 was correct 86.98\% of the time when it predicted \textit{related}, and it recovered 97.80\% of all truly related abstracts, indicating that it was very effective at not missing relevant papers.
For the unrelated class, GPT-5.4 was correct 96.70\% of the time when it predicted \textit{not related}, but its recall for that class was lower at 81.51\%, meaning that it did not label unrelated abstracts as \textit{not related} often enough.
In practical terms, GPT-5.4 functioned as a strong high-recall relevance screener: it is well suited to settings where the priority is to avoid missing relevant abstracts, but it is weaker at filtering out irrelevant abstracts cleanly.
In plain terms, the model tends to over-include, which may be acceptable for screening but increases downstream manual review effort and can introduce more false positives for later NER annotation.

\subsection{Supervised Model Details}
\label{app:supervised_model_details}

The supervised BERT model aliases and checkpoints are as follows: \texttt{biobert} = \texttt{dmis-lab/biobert-base-cased-v1.2}, \texttt{clinicalbert} = \texttt{emilyalsentzer/Bio\_ClinicalBERT}, and \texttt{pubmedbert} = \texttt{microsoft/BiomedNLP-PubMedBERT-base-}\linebreak\texttt{uncased-abstract-fulltext}.
To avoid further reducing training coverage for rare labels, we do not create a separate development split; instead, we fix the training hyperparameters in advance and apply the same settings across the BERT baselines.
No hyperparameter choices are made based on test-set results.
All BERT models use the same hyperparameters: \texttt{max\_length} = 512, \texttt{num\_train\_epochs} = 5.0, \texttt{learning\_rate} = 5e-05, \texttt{weight\_decay} = 0.01, \texttt{train\_batch\_size} = 8, \texttt{eval\_batch\_size} = 16, and \texttt{seed} = 42.
BERT flat-ner results for this 5-epoch setup are as follows.
In coarse-grained evaluation, BioBERT reaches 80.96 precision, 81.47 recall, and 81.22 F1; ClinicalBERT reaches 79.45 precision, 79.10 recall, and 79.28 F1; and PubMedBERT reaches 81.88 precision, 81.37 recall, and 81.62 F1.
In fine-grained evaluation, BioBERT reaches 78.61 precision, 78.61 recall, and 78.61 F1; ClinicalBERT reaches 78.29 precision, 78.73 recall, and 78.51 F1; and PubMedBERT reaches 78.93 precision, 80.06 recall, and 79.49 F1.
For W2NER~\cite{li2022unified}, we use PubMedBERT embeddings as the encoder input representation and train for 10 epochs.
We include it specifically because it remains supervised while also supporting the \texttt{non-flat-ner} setting, unlike the BIO-based BERT baselines.
We additionally report aggregate W2NER runs in both \texttt{flat-ner} and \texttt{non-flat-ner} settings.
In coarse-grained \texttt{flat-ner}, W2NER reaches 84.23 precision, 79.63 recall, and 81.86 F1; in fine-grained \texttt{flat-ner}, it reaches 83.07 precision, 78.01 recall, and 80.46 F1.
For \texttt{non-flat-ner}, W2NER reaches 12.40 precision, 14.81 recall, and 13.48 F1 in the coarse-grained setting, and 12.74 precision, 14.81 recall, and 13.70 F1 in the fine-grained setting.

\subsection{Zero-shot Prompting and Inference Details}
\label{app:unsupervised_prompting}

For zero-shot evaluation, we apply the GLiNER2 variants GLiNER2-short and GLiNER2-def~\cite{zaratiana2025gliner2} together with GLiNER-BioMed~\cite{yazdani2025gliner}, without supervised fine-tuning on PrionNER.
For both GLiNER2-short and GLiNER2-def, we use the checkpoint \texttt{fastino/gliner2-large-v1}; for GLiNER-BioMed, we use \texttt{Ihor/gliner-biomed-base-v1.0}.
GLiNER2-short supplies only the entity type names, whereas GLiNER2-def supplies both the entity type names and short type definitions.
GLiNER-BioMed is evaluated only with entity type names.
In our current GLiNER-based pipeline, non-flat behavior can be represented for nested and overlapping entities through multiple contiguous spans, but the model does not support explicit discontinuous multi-span entity objects.
We also tested smaller local biomedical LLMs, including \texttt{aaditya/Llama3-OpenBioLLM-8B} and \texttt{BioMistral/BioMistral-7B}, but excluded them from the main comparison because in our pipeline they frequently produced chat-style outputs, malformed or truncated JSON, corrupted labels, and non-literal spans, making them unreliable for strict exact-span NER extraction.
The models are run at the abstract level.

For the LLM-based setting, we evaluate OpenAI GPT-5.4~\cite{openai2026gpt54}, \texttt{google/gemma-4-31B-it}, and \texttt{google/gemma-4-26B-A4B-it} from the Gemma 4 family~\cite{team2024gemma} using a fixed prompt template that introduces the task, specifies the target entity schema, and requests structured outputs.
Each input instance is processed at the abstract level, and the model outputs are required to follow a constrained format such as JSON with entity text.

After inference, we apply a verification and normalization pipeline before evaluation.
For NER outputs, we first verify the output shape: the model response must be a JSON object containing an \texttt{entities} list, and for OpenAI/Gemma outputs the returned \texttt{text} field must match the input abstract after collapsing whitespace.
We then validate entity labels against the schema, rejecting invalid \texttt{coarse\_type}/\texttt{fine\_type} combinations.
In limited cases, we repair partially correct predictions: in the coarse-only schema setting, \texttt{fine\_type} is normalized to the same value as \texttt{coarse\_type}, and if a predicted \texttt{fine\_type} is valid but the \texttt{coarse\_type} is missing or incorrect, we infer the corresponding coarse label from the schema.
Next, we verify that each predicted mention can be aligned back to the source text and attach start/end offsets with deduplication.
This alignment step is necessary for the zero-shot systems because LLM- and GLiNER-style outputs typically return entity strings but not reliable character offsets, whereas exact-match NER evaluation requires position-specific spans.
When the same predicted surface form occurs multiple times in the same abstract, we map that mention to all exact string-matching occurrences in order to recover candidate offsets deterministically.
This expansion step should therefore be understood as an offset-recovery procedure for span-text outputs rather than an additional modeling component, although it introduces an evaluation asymmetry relative to supervised token-level baselines that predict positions directly.
For OpenAI/Gemma outputs, mentions that cannot be aligned are skipped and logged with warnings.

The GPT-5.4 prompt template used in our experiments is shown below.

\begin{lstlisting}[basicstyle=\ttfamily\footnotesize,breaklines=true,breakatwhitespace=false,columns=fullflexible,keepspaces=true]
System:
You are an information extraction model for biomedical case reports and reviews about prion disease.
Your job is to label entity spans using only the provided schema and output one valid JSON object.

User: Read the schema carefully and follow it exactly.

Schema JSON: {entity_schema}

Rules:
1. Extract only explicit spans from the text.
2. Use exact labels from the schema.
3. Use exact text spans from the input.
4. Do not output start or end offsets.
5. Do not output comments, markdown, or extra keys.
6. If uncertain, omit the span instead of guessing.

Input text: {text}

Output exactly one JSON object with this structure:
{
  "text": "<original input text>",
  "entities": [
    {
      "mention": "<exact span>",
      "coarse_type": "<schema coarse type>",
      "fine_type": "<schema fine type>",
      "normalized": "<normalized form or same as mention>"
    }
  ]
}
\end{lstlisting}

\section{Annotation Guidelines and Schema}
\label{sec:prionner_annotation_guidelines_short}
\label{app:annotation_guidelines}

\subsection{Guideline Principles}

Global rules: annotate only explicit mentions; prefer the most specific label available; keep spans minimal but complete; annotate both long form and abbreviation when both are explicit; treat molecular subtype strings such as \textit{MM1}, \textit{MV2}, and \textit{VV2} as \texttt{sCJD} mentions, with finer subtype interpretation handled in normalization; treat exposure or treatment phrases such as \textit{corneal transplantation}, \textit{cadaveric-derived hormone treatment} (e.g., growth hormone), and \textit{cadaveric dura mater graft} as \texttt{Treatment} mentions rather than disease mentions; include generic heads such as \textit{disease}, \textit{syndrome}, and \textit{symptom} when they are part of the explicit annotated mention; and for statistical expressions annotate only the value span (e.g., \textit{100\%}, \textit{0.06\%}) rather than surrounding cue phrases.

Do not assign labels by keyword alone.
A span is a disease mention only if it names a disease entity in context, and a span is a symptom mention only if it expresses a clinical manifestation in context.
For example, \textit{disease duration} is not a disease label, and \textit{symptom onset} is not a \texttt{Symptom} label.

\clearpage
\onecolumn
\subsection{Train Annotation Consistency and Guideline Drift}

The training annotations were produced over multiple rounds by the two annotators and then consolidated through adjudication into a single final corpus version.
Because the train split was not preserved as two separately frozen annotator-specific versions, we do not report a formal per-annotator label-distribution comparison here.
Instead, Table~\ref{tab:guideline_drift_examples} summarizes the main recurrent sources of early annotation-style drift that were resolved during guideline refinement and then applied consistently in the final schema.
Most of these changes affected boundary selection or distinctions between nearby label types rather than the overall entity inventory.

\begin{table}[H]
\centering
\small
\setlength{\tabcolsep}{4pt}
\renewcommand{\arraystretch}{1.15}
\begin{tabular}{p{0.6cm} p{5.9cm} p{7.8cm}}
\toprule
\textbf{No.} & \textbf{First Guideline / Early Practice} & \textbf{Final Guideline} \\
\midrule
1 & \textit{brain} was not labeled as \texttt{Anatomic\_location}. & \textit{brain} is labeled as \texttt{Anatomic\_location}. \\
2 & \textit{died} and \textit{death} were not labeled as \texttt{Complication}. & \textit{died} and \textit{death} are labeled as \texttt{Complication}. \\
3 & \textit{FLAIR}, \textit{T1-weighted}, \textit{T2-weighted}, \textit{ADC}, and \textit{DWI} were labeled as \texttt{Imaging\_test}. & These sequence terms are labeled as \texttt{Imaging\_sequence} rather than \texttt{Imaging\_test}. \\
4 & Time expressions such as \textit{in year 2013}, \textit{in 2013}, and \textit{12 months after onset of symptoms} were not consistently separated into anchor vs.\ duration components. & \texttt{Time\_point} captures anchor expressions such as \textit{2013} and \textit{onset of symptoms}, while \texttt{Duration} captures spans such as \textit{12 months}. \\
5 & \texttt{Autopsy} was restricted to explicit procedure mentions such as \textit{autopsy} and \textit{autopsy examination}. & \texttt{Autopsy} was expanded to include additional procedure-reporting forms such as \textit{pathologically}, \textit{autopsy}, and \textit{neuropathological examination}. \\
6 & Statistical entities were often annotated as full cue phrases, e.g., \textit{sensitivity of 61\%} or \textit{prevalence of 1/1 million people per year}. & Only the value span is annotated: e.g., \textit{61\%} as \texttt{Sensitivity} and \textit{1/1 million people per year} as \texttt{Prevalence}. \\
7 & \textit{Heidenhain variant} was not labeled as \texttt{sCJD}. & \textit{Heidenhain variant} is labeled as \texttt{sCJD}. \\
\bottomrule
\end{tabular}
\caption{Examples of guideline drift observed during training annotation and the final adjudicated guideline used in PrionNER.}
\label{tab:guideline_drift_examples}
\end{table}

\subsection{Full Entity Schema}
\small
\setlength{\tabcolsep}{3pt}
\renewcommand{\arraystretch}{1.15}
\begin{longtable}{>{\raggedright\arraybackslash}p{3.2cm} 
                   >{\raggedright\arraybackslash}p{2.9cm} 
                   >{\raggedright\arraybackslash}p{9.0cm}}
\caption{Full coarse-grained and fine-grained entity schema of PrionNER, including definitions and representative examples.}
\label{tab:coarse_fine_diagnostic_types_appendix} \\
\hline
\textbf{Coarse-grained Type} & \textbf{Fine-grained Type} & \textbf{Definitions} \\
\hline
\endfirsthead
\caption[]{Full coarse-grained and fine-grained entity schema of PrionNER, including definitions and representative examples (continued).} \\
\hline
\textbf{Coarse-grained Type} & \textbf{Fine-grained Type} & \textbf{Definitions} \\
\hline
\endhead
\hline
\multicolumn{3}{r}{Continued on next page} \\
\hline
\endfoot
\hline
\endlastfoot

\multicolumn{3}{l}{\textbf{Case Input~\cite{geschwind2015prion}}} \\
\hline
Age & Age & The age of a patient at disease onset. \newline
Examples: age 62; 45-year-old; at 58 years \\
\hline

Symptom & Symptom & Any subjective experience reported by the patient or objective observation made by a clinician. \newline
\textit{Examples:} memory loss; fatigue; hyperreflexia; Babinski sign; rigidity; startle response \\
\hline

\multirow{6}{*}{Test\_name}
& Imaging\_test & MRI; CT scan; PET scan \\
& Electrophysio\_test & nerve conduction study, Polysomnography \\
& Blood\_biomarker\_test & CSF 14-3-3 assay; tau assay; complete blood count \\
& Genetic\_test & PRNP genetic testing; mutation analysis; gene panel testing \\
& Molecular\_assay & RT-QuIC; PCR; Western blot \\
& Autopsy & brain autopsy; neuropathological examination; postmortem analysis \\
\hline

\multirow{1}{*}{Sequences}
& Imaging\_sequence & DWI; FLAIR; diffusion-weighted imaging; ADC maps \\
\hline

Anatomic\_location & Anatomic\_location &
A specific anatomical structure where a clinical finding or pathological change is observed.
\newline
\textit{Examples:} cortex; basal ganglia; thalamus; caudate nucleus; cerebral cortex \\
\hline

\multirow{2}{*}{Findings}
& Imaging\_finding & pulvinar sign; restricted diffusion; cortical ribboning \\
& Autopsy\_finding & spongiform change; neuronal loss; astrogliosis \\
\hline

\multicolumn{3}{l}{\textbf{Case Diagnosis~\cite{cdc2026prions}}} \\
\hline
Generic\_Prion
& Generic\_Prion &
Prion disease mentions without specific subtype classification.
\newline
\textit{Examples:} prion disease; CJD, bovine spongiform encephalopathy, BSE, mad cow disease, Transmissible Spongiform Encephalopathy, TSE \\
\hline

\multirow{3}{*}{Sporadic\_Prion}
& sCJD &
Sporadic Creutzfeldt-Jakob Disease \\
& sFI &
Sporadic Fatal Insomnia \\
& VPSPr &
Variably Protease-Sensitive Prionopathy \\
\hline

\multirow{3}{*}{Familial\_Prion}
& fCJD &
Familial Creutzfeldt-Jakob Disease \\
& GSS &
Gerstmann-Sträussler-Scheinker Syndrome \\
& FFI &
Fatal Familial Insomnia \\
\hline

\multirow{3}{*}{Acquired\_Prion}
& vCJD &
Variant Creutzfeldt-Jakob Disease \\
& iCJD &
Iatrogenic Creutzfeldt-Jakob Disease \\
& Kuru &
Kuru \\
\hline

Differential\_Diagnosis & Differential\_Diagnosis &
Non-prion diseases used for differential diagnosis (no fine-grained subtypes in this schema).
\newline
\textit{Examples:} Alzheimer’s disease; autoimmune encephalitis; viral encephalitis; Parkinson’s disease \\
\hline

\multicolumn{3}{l}{\textbf{Clinical Course and Context}} \\
\hline
Treatment & Treatment &
A therapeutic intervention, medication used to manage a disease or its symptoms.
\newline
Examples: supportive care; quinacrine; doxycycline; symptomatic treatment~\cite{benavente2024therapeutic} \\
\hline

Complication & Complication &
A secondary medical condition that arises as a consequence of disease progression.
\newline
Examples: pneumonia; aspiration pneumonia; respiratory failure~\cite{willis2006medical} \\
\hline

\multirow{2}{*}{Time}
& Duration &
A span or length of time over which a clinical event, symptom, or disease progression occurs.
\newline
\textit{Examples:} within 3 months; over 2 years; rapidly progressive over weeks \\
& Time\_point &
A specific point in time associated with clinical symptoms or a specific event.
\newline
\textit{Examples:} at onset; at age 62; in 2021 \\
\hline

\multirow{4}{*}{Stats}
& Sensitivity &
The true positive rate for a diagnostic test~\cite{saltelli2019so}.
\newline
\textit{Examples:} sensitivity of 85\% \\
& Specificity &
The true negative rate for a diagnostic test~\cite{saltelli2019so}.
\newline
\textit{Examples:} specificity of 92\% \\
& Prevalence &
Disease frequency within a population~\cite{tenny2017prevalence}.
\newline
\textit{Examples:} 1 per million; prevalence of 0.5\% \\
& Incidence &
The rate at which new cases of a disease
occur in a population during a specified time period.
\newline
\textit{Examples:} annual incidence of 2 per million \\
\hline
\end{longtable}
\renewcommand{\arraystretch}{1}

\subsection{Compact Annotation Guideline}

\footnotesize
\setlength{\tabcolsep}{4pt}
\renewcommand{\arraystretch}{1.15}
\newcommand{\guidelinelabel}[1]{\shortstack[l]{#1}}
\begin{longtable}{>{\raggedright\arraybackslash}p{2.2cm}
>{\raggedright\arraybackslash}p{3.0cm}
>{\raggedright\arraybackslash}p{4.2cm}
>{\raggedright\arraybackslash}p{4.8cm}}
\caption{Compact PrionNER annotation guideline.}
\label{tab:prionner_annotation_guideline_short_table} \\
\hline
\textbf{Label} & \textbf{Use For} & \textbf{Annotate} & \textbf{Do Not Annotate} \\
\hline
\endfirsthead
\hline
\textbf{Label} & \textbf{Use For} & \textbf{Annotate} & \textbf{Do Not Annotate} \\
\hline
\endhead
\hline
\endfoot

\guidelinelabel{\texttt{Generic\_}\\\texttt{Prion}} &
Prion disease mentions without explicit subtype classification &
\textit{prion disease}, \textit{prion diseases}, \textit{Creutzfeldt-Jakob disease}, \textit{CJD}, \textit{bovine spongiform encephalopathy}, \textit{BSE}, \textit{mad cow disease} &
Specific subtype mentions such as \textit{sporadic CJD} or \textit{variant CJD}; protein mentions such as \textit{prion protein}, \textit{PrP}, or \textit{PrP\textsuperscript{Sc}}; pathological finding uses such as \textit{spongiform encephalopathy} when used as an autopsy finding \\
\hline

\texttt{sCJD} &
Sporadic Creutzfeldt-Jakob disease &
\textit{sporadic Creutzfeldt-Jakob disease}, \textit{sporadic CJD}, \textit{sCJD}, \textit{MM1}, \textit{MV1}, \textit{VV2} when clearly used as sCJD subtypes &
Generic \textit{CJD} without a sporadic cue \\
\hline

\texttt{sFI} &
Sporadic fatal insomnia &
\textit{sporadic fatal insomnia}, \textit{sFI} &
\textit{fatal familial insomnia}, \textit{FFI}, \textit{FFI-1}, \textit{FFI-2} \\
\hline

\texttt{VPSPr} &
Variably protease-sensitive prionopathy &
\textit{variably protease-sensitive prionopathy}, \textit{VPSPr} &
\textit{PrP}, \textit{PrP\textsuperscript{Sc}}, \textit{PrP-res}, or other protein-level mentions \\
\hline

\texttt{fCJD} &
Familial or hereditary CJD &
\textit{familial CJD}, \textit{familial Creutzfeldt-Jakob disease}, \textit{hereditary CJD}, \textit{genetic CJD}, \textit{fCJD} &
Mutation or genotype mentions alone \\
\hline

\texttt{GSS} &
Gerstmann-Str\"aussler-Scheinker syndrome &
\textit{Gerstmann-Str\"aussler-Scheinker syndrome}, \textit{GSS}, \textit{GSS102}, \textit{GSS105} &
Mutation names alone; pathological descriptors such as \textit{kuru plaques} \\
\hline

\texttt{FFI} &
Fatal familial insomnia &
\textit{fatal familial insomnia}, \textit{FFI}, \textit{FFI-1}, \textit{FFI-2} &
Isolated \textit{insomnia}; use \texttt{Symptom} instead \\
\hline

\texttt{vCJD} &
Variant CJD &
\textit{variant Creutzfeldt-Jakob disease}, \textit{variant CJD}, \textit{vCJD}, \textit{new variant CJD}, \textit{nvCJD} &
Generic uses of \textit{variant} without clear disease reference \\
\hline

\texttt{iCJD} &
Iatrogenic CJD &
\textit{iatrogenic Creutzfeldt-Jakob disease}, \textit{iatrogenic CJD}, \textit{growth hormone-associated CJD}, \textit{dural graft associated CJD}, \textit{iCJD}, \textit{dCJD} &
\textit{iatrogenic} alone when the CJD referent is unclear \\
\hline

\texttt{Kuru} &
Kuru disease mention &
\textit{Kuru}, \textit{kuru} &
\textit{kuru plaques}; use \texttt{Autopsy\_finding} instead \\
\hline

\guidelinelabel{\texttt{Differential\_}\\\texttt{Diagnosis}} &
Non-prion diseases considered as alternatives or comparators &
\textit{Alzheimer disease}, \textit{autoimmune encephalitis}, \textit{Huntington disease} &
Prion diseases; symptom phrases unless they are clearly used as alternative diagnoses \\
\hline

\texttt{Symptom} &
Symptoms, signs, and clinical manifestations &
\textit{dementia}, \textit{myoclonus}, \textit{cerebellar ataxia}, \textit{psychotic symptoms}, \textit{hyperreflexia} &
Disease names; imaging findings; autopsy findings \\
\hline

\guidelinelabel{\texttt{Imaging\_}\\\texttt{test}} &
Imaging modalities or procedures &
\textit{MRI}, \textit{CT}, \textit{PET}, \textit{SPECT} &
Sequences such as \textit{FLAIR}, \textit{DWI}; findings such as \textit{pulvinar sign} \\
\hline

\guidelinelabel{\texttt{Electrophysio\_}\\\texttt{test}} &
Electrophysiology procedures &
\textit{EEG}, \textit{electroencephalogram}, \textit{polysomnography} &
Electrophysiologic findings or waveform targets \\
\hline

\guidelinelabel{\texttt{Blood\_}\\\texttt{biomarker\_}\\\texttt{test}} &
Biomarker assays and specimen-as-test shorthand &
\textit{CSF}, \textit{tau assay}, \textit{blood tests} &
Molecular methods such as \textit{PCR}, \textit{RT-QuIC}, or \textit{Western blot} \\
\hline

\guidelinelabel{\texttt{Genetic\_}\\\texttt{test}} &
Genetic testing procedures &
\textit{genetic testing}, \textit{Prion gene analysis}, \textit{genetic analysis} &
Gene names and mutation names alone \\
\hline

\guidelinelabel{\texttt{Molecular\_}\\\texttt{assay}} &
Molecular or biochemical assays &
\textit{RT-QuIC}, \textit{PCR}, \textit{Western blot}, \textit{molecular analysis} &
Biopsy or autopsy procedures; gene targets without a testing method \\
\hline

\texttt{Autopsy} &
Postmortem or tissue examination procedures &
\textit{brain autopsy}, \textit{postmortem examination}, \textit{brain biopsy} &
Pathological findings themselves; anatomy alone \\
\hline

\guidelinelabel{\texttt{Imaging\_}\\\texttt{sequence}} &
Acquisition or sequence terms &
\textit{DWI}, \textit{FLAIR}, \textit{T2 weighted}, \textit{ADC maps} &
Imaging modality names; imaging abnormalities \\
\hline

\guidelinelabel{\texttt{Imaging\_}\\\texttt{finding}} &
Radiologic abnormalities or named signs &
\textit{pulvinar sign}, \textit{restricted diffusion}, \textit{cortical ribboning} &
Test modality alone; anatomy alone unless separately annotated \\
\hline

\guidelinelabel{\texttt{Autopsy\_}\\\texttt{finding}} &
Pathological or postmortem findings &
\textit{spongiform change}, \textit{neuronal loss}, \textit{gliosis}, \textit{florid plaques}, \textit{kuru plaques}, \textit{spongiform encephalopathy} when used as a pathological finding &
Procedure terms; disease names \\
\hline

\guidelinelabel{\texttt{Anatomic\_}\\\texttt{location}} &
Body locations linked to findings or symptoms &
\textit{thalamus}, \textit{basal ganglia}, \textit{caudate nucleus}, \textit{cortex}, \textit{deep white matter} &
Non-anatomical descriptors; whole finding phrase when only one part is anatomy \\
\hline

\texttt{Age} &
Age or age-at-onset expression &
\textit{58-year-old}, \textit{age 62}, \textit{young adults} &
Durations; calendar dates \\
\hline

\texttt{Treatment} &
Therapies, medications, or care strategies &
\textit{palliative care}, \textit{quinacrine}, \textit{doxycycline}, \textit{antipsychotics}; annotate the full treatment mention when expressed as a therapy or care strategy &
Diagnostic tests; non-intervention goals unless expressed as actual therapy \\
\hline

\texttt{Complication} &
Secondary adverse conditions or end-stage outcomes &
\textit{pneumonia}, \textit{respiratory failure}, \textit{SIADH}, \textit{death}, \textit{died} &
The primary prion disease itself \\
\hline

\texttt{Duration} &
Time span or length &
\textit{within 3 months}, \textit{over 2 years}, \textit{13 months} &
Calendar dates; age expressions \\
\hline

\guidelinelabel{\texttt{Time\_}\\\texttt{point}} &
Specific time anchor &
\textit{at onset}, \textit{on admission}, \textit{1996}, \textit{January 2012} &
Durations; exclude discourse connectives such as \textit{in}, \textit{after} when they are not themselves the time anchor \\
\hline

\texttt{Sensitivity} &
Diagnostic sensitivity expression &
\textit{100\%}, \textit{91\%} when they are explicitly the sensitivity value, e.g., annotate only \textit{100\%} in \textit{sensitivity of 100\%} &
The cue word alone, e.g., \textit{sensitivity}, when no value is included; unrelated percentages \\
\hline

\texttt{Specificity} &
Diagnostic specificity expression &
\textit{92\%}, \textit{95\%} when they are explicitly the specificity value &
The cue word alone, e.g., \textit{specificity}, when no value is included; unrelated percentages \\
\hline

\texttt{Prevalence} &
Disease frequency expression &
\textit{1--2 people per million annually}, \textit{0.5\%} when it is explicitly the prevalence value &
The cue word alone, e.g., \textit{prevalence}, when no value is included; sample size counts \\
\hline

\texttt{Incidence} &
New-case rate expression &
\textit{0.06\%}, \textit{2 per million annually}, \textit{0.37 cases/million}, \textit{1 in 1 000 000} when they are explicitly the incidence value &
The cue word alone, e.g., \textit{incidence}, when no value is included; sample size counts; surrounding time or cue phrases such as \textit{in 2023}, \textit{million per year}, or \textit{annual incidence of} \\
\hline
\end{longtable}

\normalsize \FloatBarrier
\clearpage
\twocolumn

\section{Extended Dataset Statistics}
\label{app:additional_dataset_statistics}

\subsection{Structural Annotation Statistics}

The train and test splits both contain non-trivial structural complexity at the span level (Table~\ref{tab:test_structural_statistics}).
Discontinuous entities remain relatively uncommon in absolute terms, but they appear in both splits and are distributed across clinically salient labels rather than being confined to a single category.
In both train and test, \textit{Symptom} is the most common discontinuous label, followed by \textit{Anatomic\_location}; train also shows notable discontinuous cases for \textit{Imaging\_finding}, while test includes comparatively more \textit{Blood\_biomarker\_test} discontinuities (Table~\ref{tab:discontinuous_entity_statistics}).

Nested structures are much more frequent than overlapping ones, and they are concentrated in recurring clinically meaningful label pairs (Tables~\ref{tab:nested_entity_statistics} and~\ref{tab:overlapping_entity_statistics}).
In the training split, the most common nested envelope patterns are \textit{Symptom} within \textit{Symptom}, \textit{Duration} with \textit{Time\_point}, and \textit{Anatomic\_location} paired with \textit{Symptom}; in the test split, the dominant patterns shift toward \textit{Anatomic\_location} with \textit{Symptom} or \textit{Imaging\_finding}, with an additional cluster of \textit{Treatment}--\textit{iCJD} nesting.
By contrast, overlapping envelope pairs are rare in both splits, suggesting that the benchmark's structural difficulty is driven primarily by discontinuity and especially nesting rather than widespread partial-overlap phenomena.

\begin{table}[H]
\centering
\small
\setlength{\tabcolsep}{3pt}
\resizebox{\columnwidth}{!}{%
\begin{tabular}{lrrrrr}
\toprule
\textbf{Split} & \textbf{Docs} & \textbf{Entities} & \textbf{Disc.} & \textbf{Nested Env.} & \textbf{Overlap Env.} \\
\midrule
Train & 247 & 5,149 & 97 & 235 & 5 \\
Test & 70 & 1,806 & 34 & 80 & 2 \\
\bottomrule
\end{tabular}
}
\caption{Span-structure summary for the train and test splits.
Discontinuous entities are text-bound entities with more than one atomic span.
Nested and overlapping pair counts are reported at the entity-envelope level.}
\label{tab:test_structural_statistics}
\end{table}

\begin{table}[H]
\centering
\small
\setlength{\tabcolsep}{4pt}
\begin{tabular}{lrr}
\toprule
\textbf{Label} & \textbf{Train} & \textbf{Test} \\
\midrule
Symptom & 36 & 10 \\
Anatomic\_location & 15 & 9 \\
Imaging\_finding & 14 & 3 \\
Blood\_biomarker\_test & 4 & 5 \\
Differential\_Diagnosis & 4 & 2 \\
Treatment & 4 & 1 \\
Duration & 3 & 2 \\
Imaging\_sequence & 4 & 0 \\
Imaging\_test & 4 & 0 \\
Autopsy\_finding & 3 & 1 \\
sCJD & 3 & 0 \\
vCJD & 1 & 1 \\
fCJD & 1 & 0 \\
iCJD & 1 & 0 \\
\bottomrule
\end{tabular}
\caption{Discontinuous entities by label in the train and test splits.
The train split contains discontinuous entities in 54 documents, and the test split contains discontinuous entities in 20 documents.}
\label{tab:discontinuous_entity_statistics}
\end{table}

\begin{table}[H]
\centering
\footnotesize
\setlength{\tabcolsep}{4pt}
\begin{tabular}{>{\raggedright\arraybackslash}p{0.30\columnwidth}>{\raggedright\arraybackslash}p{0.30\columnwidth}rr}
\toprule
\textbf{Label A} & \textbf{Label B} & \textbf{Train} & \textbf{Test} \\
\midrule
Symptom & Symptom & 42 & 9 \\
\shortstack[l]{Anatomic\_\\location} & Symptom & 28 & 13 \\
Duration & \shortstack[l]{Time\_\\point} & 30 & 2 \\
\shortstack[l]{Anatomic\_\\location} & \shortstack[l]{Anatomic\_\\location} & 18 & 10 \\
\shortstack[l]{Anatomic\_\\location} & \shortstack[l]{Imaging\_\\finding} & 14 & 11 \\
\shortstack[l]{Anatomic\_\\location} & \shortstack[l]{Autopsy\_\\finding} & 16 & 4 \\
\shortstack[l]{Imaging\_\\finding} & \shortstack[l]{Imaging\_\\sequence} & 9 & 2 \\
\shortstack[l]{Anatomic\_\\location} & \shortstack[l]{Differential\_\\Diagnosis} & 8 & 0 \\
\shortstack[l]{Blood\_\\biomarker\_\\test} & \shortstack[l]{Blood\_\\biomarker\_\\test} & 4 & 4 \\
Treatment & iCJD & 1 & 6 \\
\shortstack[l]{Imaging\_\\sequence} & \shortstack[l]{Imaging\_\\sequence} & 6 & 0 \\
\shortstack[l]{Differential\_\\Diagnosis} & \shortstack[l]{Differential\_\\Diagnosis} & 4 & 2 \\
\shortstack[l]{Generic\_\\Prion} & \shortstack[l]{Time\_\\point} & 4 & 2 \\
Duration & Duration & 2 & 3 \\
\shortstack[l]{Imaging\_\\finding} & \shortstack[l]{Imaging\_\\finding} & 4 & 0 \\
\shortstack[l]{Autopsy\_\\finding} & Kuru & 3 & 1 \\
\shortstack[l]{Anatomic\_\\location} & Treatment & 3 & 0 \\
\shortstack[l]{Imaging\_\\sequence} & \shortstack[l]{Imaging\_\\test} & 3 & 0 \\
\shortstack[l]{Differential\_\\Diagnosis} & Symptom & 2 & 1 \\
\shortstack[l]{Blood\_\\biomarker\_\\test} & \shortstack[l]{Differential\_\\Diagnosis} & 0 & 3 \\
\bottomrule
\end{tabular}
\caption{Nested entity-envelope pairs by label pair in the train and test splits.}
\label{tab:nested_entity_statistics}
\end{table}

\begin{table}[H]
\centering
\footnotesize
\setlength{\tabcolsep}{4pt}
\begin{tabular}{>{\raggedright\arraybackslash}p{0.30\columnwidth}>{\raggedright\arraybackslash}p{0.30\columnwidth}rr}
\toprule
\textbf{Label A} & \textbf{Label B} & \textbf{Train} & \textbf{Test} \\
\midrule
\shortstack[l]{Anatomic\_\\location} & \shortstack[l]{Anatomic\_\\location} & 1 & 0 \\
\shortstack[l]{Autopsy\_\\finding} & \shortstack[l]{Generic\_\\Prion} & 1 & 0 \\
FFI & sFI & 1 & 0 \\
\shortstack[l]{Imaging\_\\test} & \shortstack[l]{Imaging\_\\test} & 1 & 0 \\
Symptom & Symptom & 1 & 0 \\
\shortstack[l]{Blood\_\\biomarker\_\\test} & \shortstack[l]{Differential\_\\Diagnosis} & 0 & 1 \\
\shortstack[l]{Time\_\\point} & Treatment & 0 & 1 \\
\bottomrule
\end{tabular}
\caption{Overlapping entity-envelope pairs by label pair in the train and test splits.}
\label{tab:overlapping_entity_statistics}
\end{table}

\subsection{Split-Level Entity Summary}

\setlength{\tabcolsep}{5pt}
\begin{table}[t]
\centering
\small
\resizebox{\columnwidth}{!}{%
\begin{tabular}{lrrrr}
\toprule
\textbf{Split} & \textbf{Abstracts} & \textbf{Entity Types} & \textbf{Mentions} & \textbf{Unique Surface Forms} \\
\midrule
Train & 247 & 28 & 4,655 & 1,889 \\
Test & 70 & 29 & 1,650 & 729 \\
\bottomrule
\end{tabular}
}
\caption{Split-level summary of schema-defined non-meta entity annotations and unique normalized surface forms.}
\label{tab:split_level_entity_summary}
\end{table}

\subsection{Entity Surface-Form Dictionary Summary}

Unique surface forms are counted per entity type after lowercasing, trimming edge whitespace, collapsing internal whitespace, and joining discontinuous spans with spaces.

\setlength{\tabcolsep}{4pt}
\begin{table}[t]
\centering
\small
\resizebox{\columnwidth}{!}{%
\begin{tabular}{>{\raggedright\arraybackslash}p{0.34\linewidth}rrrr}
\toprule
\textbf{Entity Type} & \textbf{Train Unique} & \textbf{Test Unique} & \textbf{Train Mentions} & \textbf{Test Mentions} \\
\midrule
Symptom & 424 & 148 & 791 & 248 \\
Anatomic\_location & 212 & 89 & 646 & 223 \\
Duration & 163 & 41 & 217 & 49 \\
Time\_point & 127 & 39 & 152 & 47 \\
Treatment & 102 & 59 & 126 & 88 \\
Autopsy\_finding & 110 & 32 & 215 & 42 \\
\shortstack[l]{Differential\_\\Diagnosis} & 113 & 27 & 188 & 36 \\
Age & 108 & 25 & 137 & 29 \\
Imaging\_finding & 92 & 38 & 153 & 71 \\
Imaging\_test & 72 & 22 & 261 & 75 \\
Generic\_Prion & 61 & 27 & 741 & 233 \\
Autopsy & 54 & 25 & 199 & 64 \\
Imaging\_sequence & 38 & 27 & 117 & 77 \\
Electrophysio\_test & 42 & 10 & 173 & 38 \\
Molecular\_assay & 20 & 29 & 23 & 45 \\
\shortstack[l]{Blood\_\\biomarker\_\\test} & 23 & 14 & 64 & 39 \\
sCJD & 18 & 16 & 86 & 80 \\
iCJD & 18 & 10 & 32 & 16 \\
Complication & 18 & 6 & 78 & 19 \\
GSS & 14 & 8 & 32 & 16 \\
\bottomrule
\end{tabular}
}
\caption{Train/test comparison of unique normalized surface forms and non-meta mention counts for the most frequent entity types.}
\label{tab:entity-surface-dictionary-summary}
\end{table}

\section{Extended Results}

\subsection{Overall Performance}

\begin{figure}[t]
\centering
\includegraphics[width=\linewidth]{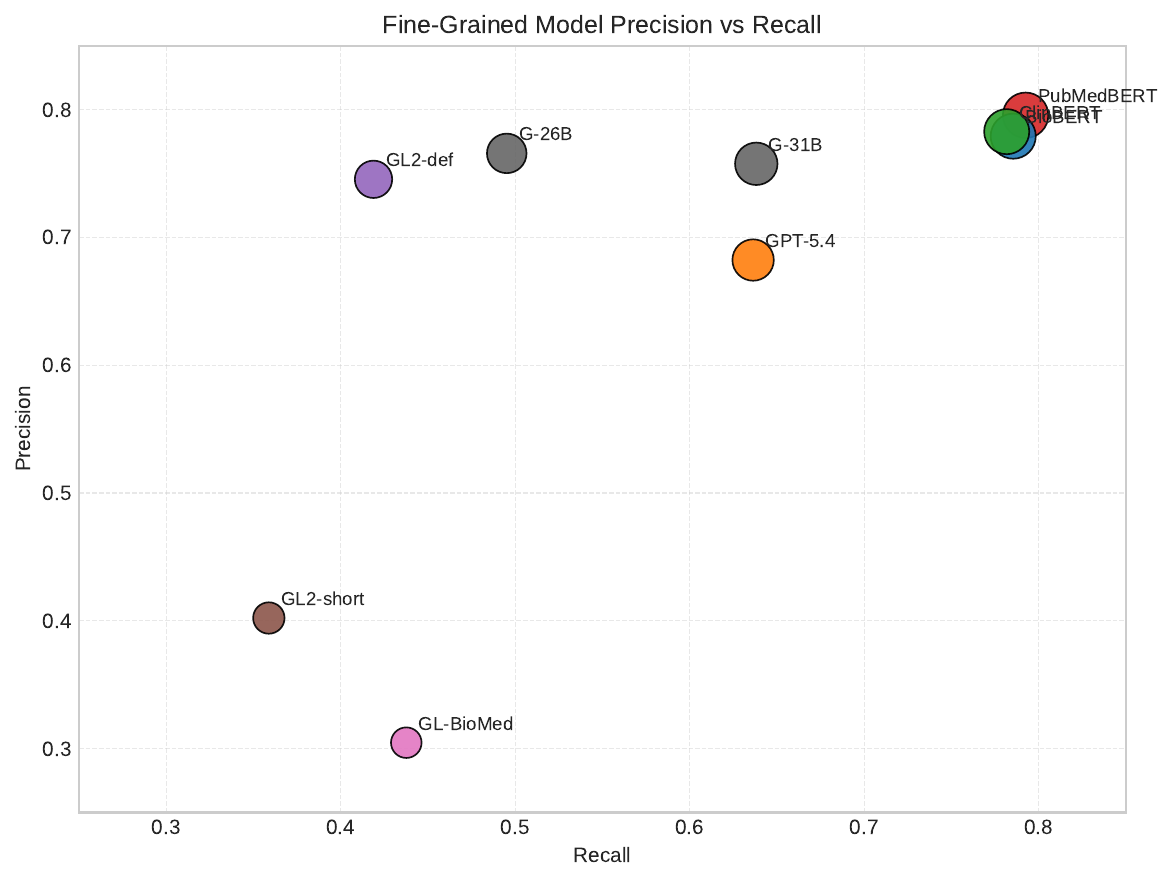}
\caption{Precision--recall trade-offs for fine-grained entity extraction across the evaluated models.}
\label{fig:fine_precision_recall_scatter}
\end{figure}

Figure~\ref{fig:fine_precision_recall_scatter} complements the main fine-grained results by showing the precision--recall trade-offs across the evaluated models.
The supervised BERT baselines cluster in the strongest region overall, while the zero-shot models show a wider spread and sharper trade-offs; among them, Gemma-4-31B occupies the strongest overall position in the fine-grained setting.

\subsection{Per-Type Fine-grained Results}

\begin{figure*}[t]
\centering
\includegraphics[width=\textwidth]{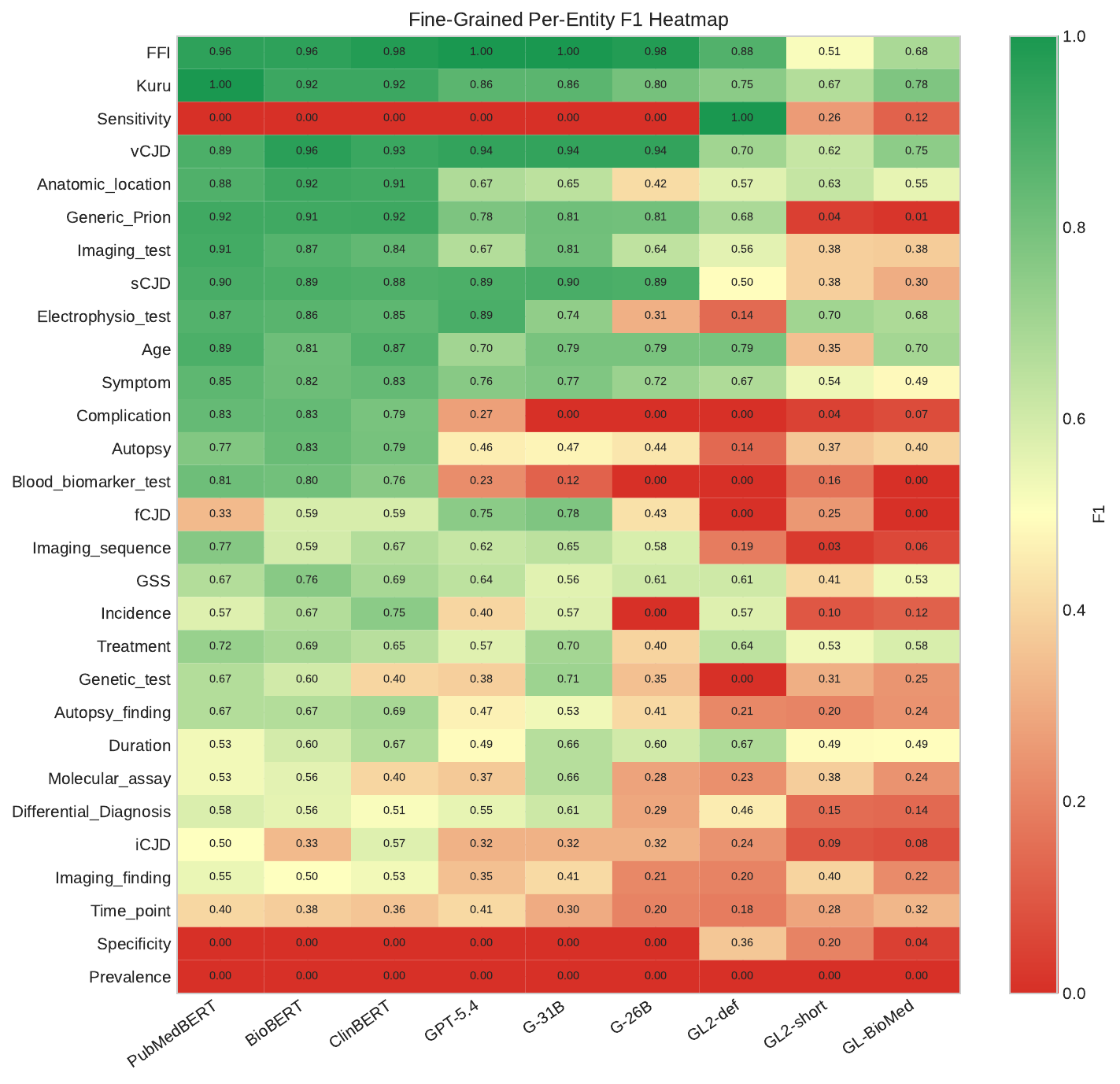}
\caption{Per-type fine-grained F1 heatmap for the main supervised and zero-shot models on the test set.}
\label{fig:fine_per_entity_f1_heatmap}
\end{figure*}

Figure~\ref{fig:fine_per_entity_f1_heatmap} and Table~\ref{tab:per_type_results_main} show that fine-grained performance is highly uneven across the schema.
The heatmap makes clear that there is no single difficulty level for PrionNER: some labels are solved reliably by most strong systems, while others remain unstable even for the best models.
This heterogeneity tracks both frequency and semantic specificity.
Common and lexically anchored categories such as \textit{Generic\_Prion}, \textit{sCJD}, \textit{Imaging\_test}, \textit{Symptom}, \textit{Age}, and \textit{Anatomic\_location} are comparatively robust, whereas sparse or context-dependent labels such as \textit{Prevalence}, \textit{Specificity}, \textit{Sensitivity}, \textit{Time\_point}, \textit{Molecular\_assay}, and subtype distinctions such as \textit{iCJD} and \textit{fCJD} remain much harder.

The supervised encoders occupy the strongest region overall and are more consistent across label families.
PubMedBERT in particular is among the top performers on many of the clinically common labels, reaching 91.77 on \textit{Generic\_Prion}, 90.79 on \textit{Imaging\_test}, 89.74 on \textit{sCJD}, 88.89 on \textit{Age}, 84.95 on \textit{Symptom}, and 81.48 on \textit{Blood\_biomarker\_test}.
BioBERT and ClinicalBERT show a similar profile and outperform PubMedBERT on a few categories, such as \textit{Anatomic\_location}, \textit{vCJD}, \textit{Complication}, \textit{Autopsy}, \textit{Autopsy\_finding}, \textit{Incidence}, and \textit{iCJD}.
Taken together, the supervised models suggest that once training data are available, the main gains come from broad coverage across the entire schema rather than isolated wins on a handful of labels.

The zero-shot models show a much sharper specialization pattern.
Gemma-4-31B is the strongest zero-shot model overall, but its strengths are concentrated in a narrower subset of distinctive labels, including \textit{FFI} (100.00), \textit{fCJD} (77.78), \textit{Genetic\_test} (71.43), \textit{Molecular\_assay} (65.79), \textit{Duration} (65.98), and \textit{Differential\_Diagnosis} (61.29).
GPT-5.4 is competitive on several labels and is best on \textit{Electrophysio\_test} (89.19) and \textit{Time\_point} (40.78), while GLiNER2 occasionally produces isolated best scores on very sparse categories such as \textit{Sensitivity}, \textit{Specificity}, and \textit{Duration}.
However, these zero-shot wins are scattered and do not translate into the same schema-wide stability seen in the supervised baselines.

Two further patterns are worth noting.
First, the imaging and diagnostic-context families remain difficult even when overall F1 is moderate: \textit{Imaging\_finding}, \textit{Imaging\_sequence}, \textit{Blood\_biomarker\_test}, \textit{Autopsy\_finding}, and \textit{Time\_point} all show larger cross-model variation than canonical disease-name labels.
Second, the most extreme values should be interpreted with caution for very rare labels.
For example, \textit{Prevalence} remains at 0.00 for all models, and isolated perfect or near-perfect results on labels such as \textit{Sensitivity} or \textit{FFI} reflect very small test counts rather than uniformly solved clinical reasoning.
Overall, the per-type view reinforces the main conclusion of the paper: PrionNER rewards models that combine strong lexical grounding on common biomedical entities with fine-grained contextual discrimination on rarer, semantically adjacent categories.

\begin{table*}[t]
\centering
\scriptsize
\setlength{\tabcolsep}{3pt}
\resizebox{\textwidth}{!}{%
\begin{tabular}{lrrr|rrrrrr}
\toprule
\multirow{2}{*}{\textbf{Entity Type}} & \multicolumn{3}{c|}{\textbf{Supervised}} & \multicolumn{6}{c}{\textbf{Zero-shot}} \\
\cmidrule(lr){2-4} \cmidrule(lr){5-10}
 & \textbf{PubMedBERT} & \textbf{BioBERT} & \textbf{ClinBERT} & \textbf{GPT-5.4} & \textbf{G-31B} & \textbf{G-26B} & \textbf{GL2-def} & \textbf{GL2-short} & \textbf{GL-BioMed} \\
\midrule
FFI & 95.65 & 95.65 & 97.78 & \textbf{100.00} & \textbf{100.00} & 97.78 & 87.80 & 51.43 & 68.42 \\
Kuru & \textbf{100.00} & 92.31 & 92.31 & 85.71 & 85.71 & 80.00 & 75.00 & 66.67 & 77.78 \\
Sensitivity & 0.00 & 0.00 & 0.00 & 0.00 & 0.00 & 0.00 & \textbf{100.00} & 25.81 & 12.24 \\
vCJD & 88.70 & \textbf{96.23} & 92.73 & 94.34 & 94.34 & 94.34 & 70.37 & 62.26 & 74.77 \\
Anatomic\_location & 88.28 & \textbf{91.88} & 91.18 & 67.19 & 64.79 & 41.61 & 56.81 & 62.83 & 55.35 \\
Generic\_Prion & 91.77 & 91.45 & \textbf{91.81} & 78.38 & 80.95 & 80.69 & 68.33 & 3.86 & 1.35 \\
Imaging\_test & \textbf{90.79} & 86.96 & 84.34 & 66.67 & 80.50 & 63.83 & 55.93 & 38.37 & 37.70 \\
sCJD & \textbf{89.74} & 89.03 & 87.90 & 88.89 & 89.66 & 88.89 & 49.54 & 38.37 & 30.19 \\
Electrophysio\_test & 87.18 & 86.08 & 85.00 & \textbf{89.19} & 73.85 & 31.11 & 14.29 & 70.42 & 67.74 \\
Age & \textbf{88.89} & 81.48 & 86.79 & 70.18 & 79.25 & 79.25 & 79.25 & 34.92 & 69.84 \\
Symptom & \textbf{84.95} & 81.97 & 83.21 & 76.16 & 77.41 & 71.86 & 67.28 & 53.69 & 48.72 \\
Complication & \textbf{83.33} & \textbf{83.33} & 78.95 & 27.27 & 0.00 & 0.00 & 0.00 & 4.40 & 6.90 \\
Autopsy & 77.31 & \textbf{82.76} & 79.31 & 45.83 & 47.42 & 43.90 & 13.89 & 36.54 & 39.66 \\
Blood\_biomarker\_test & \textbf{81.48} & 80.49 & 75.86 & 22.64 & 12.20 & 0.00 & 0.00 & 16.33 & 0.00 \\
fCJD & 33.33 & 58.82 & 58.82 & 75.00 & \textbf{77.78} & 42.86 & 0.00 & 25.00 & 0.00 \\
Imaging\_sequence & \textbf{76.51} & 59.42 & 66.67 & 62.30 & 64.62 & 58.33 & 18.60 & 2.50 & 6.20 \\
GSS & 66.67 & \textbf{75.86} & 68.97 & 64.29 & 56.25 & 60.61 & 60.87 & 40.91 & 53.33 \\
Incidence & 57.14 & 66.67 & \textbf{75.00} & 40.00 & 57.14 & 0.00 & 57.14 & 9.52 & 11.76 \\
Treatment & \textbf{72.39} & 69.01 & 65.43 & 56.94 & 69.57 & 39.64 & 64.43 & 53.27 & 58.45 \\
Genetic\_test & 66.67 & 60.00 & 40.00 & 38.46 & \textbf{71.43} & 34.78 & 0.00 & 30.77 & 24.56 \\
Autopsy\_finding & 66.67 & 66.67 & \textbf{68.82} & 46.67 & 53.12 & 40.68 & 21.28 & 20.34 & 24.18 \\
Duration & 53.06 & 59.57 & 66.67 & 49.02 & 65.98 & 59.77 & \textbf{67.35} & 49.02 & 49.48 \\
Molecular\_assay & 52.78 & 56.00 & 40.00 & 36.92 & \textbf{65.79} & 28.07 & 23.08 & 38.20 & 23.91 \\
Differential\_Diagnosis & 57.97 & 55.56 & 51.43 & 55.17 & \textbf{61.29} & 28.57 & 45.61 & 14.81 & 13.79 \\
iCJD & 50.00 & 33.33 & \textbf{57.14} & 31.58 & 31.58 & 31.58 & 24.00 & 9.23 & 7.69 \\
Imaging\_finding & \textbf{54.55} & 50.38 & 53.03 & 34.92 & 41.18 & 21.43 & 20.25 & 39.71 & 21.90 \\
Time\_point & 40.43 & 38.10 & 36.00 & \textbf{40.78} & 29.85 & 20.00 & 18.46 & 28.32 & 32.43 \\
Specificity & 0.00 & 0.00 & 0.00 & 0.00 & 0.00 & 0.00 & \textbf{36.36} & 20.00 & 3.92 \\
Prevalence & \textbf{0.00} & \textbf{0.00} & \textbf{0.00} & \textbf{0.00} & \textbf{0.00} & \textbf{0.00} & \textbf{0.00} & \textbf{0.00} & \textbf{0.00} \\
\bottomrule
\end{tabular}
}
\caption{Per-type fine-grained F1 scores for selected supervised and zero-shot models on the test set, sorted by the best score achieved for each entity type.
Header abbreviations: ClinBERT = ClinicalBERT; GL2-def = GLiNER2-def; GL2-short = GLiNER2-short; GL-BioMed = GLiNER-BioMed; G-26B and G-31B = Gemma 4 26B and 31B.}
\label{tab:per_type_results_main}
\end{table*}

\subsection{Entity-only Confusion Analysis}
\label{app:entity_only_confusion_analysis}

\begin{figure*}[t]
\centering
\begin{subfigure}[t]{0.82\textwidth}
\centering
\includegraphics[width=\linewidth]{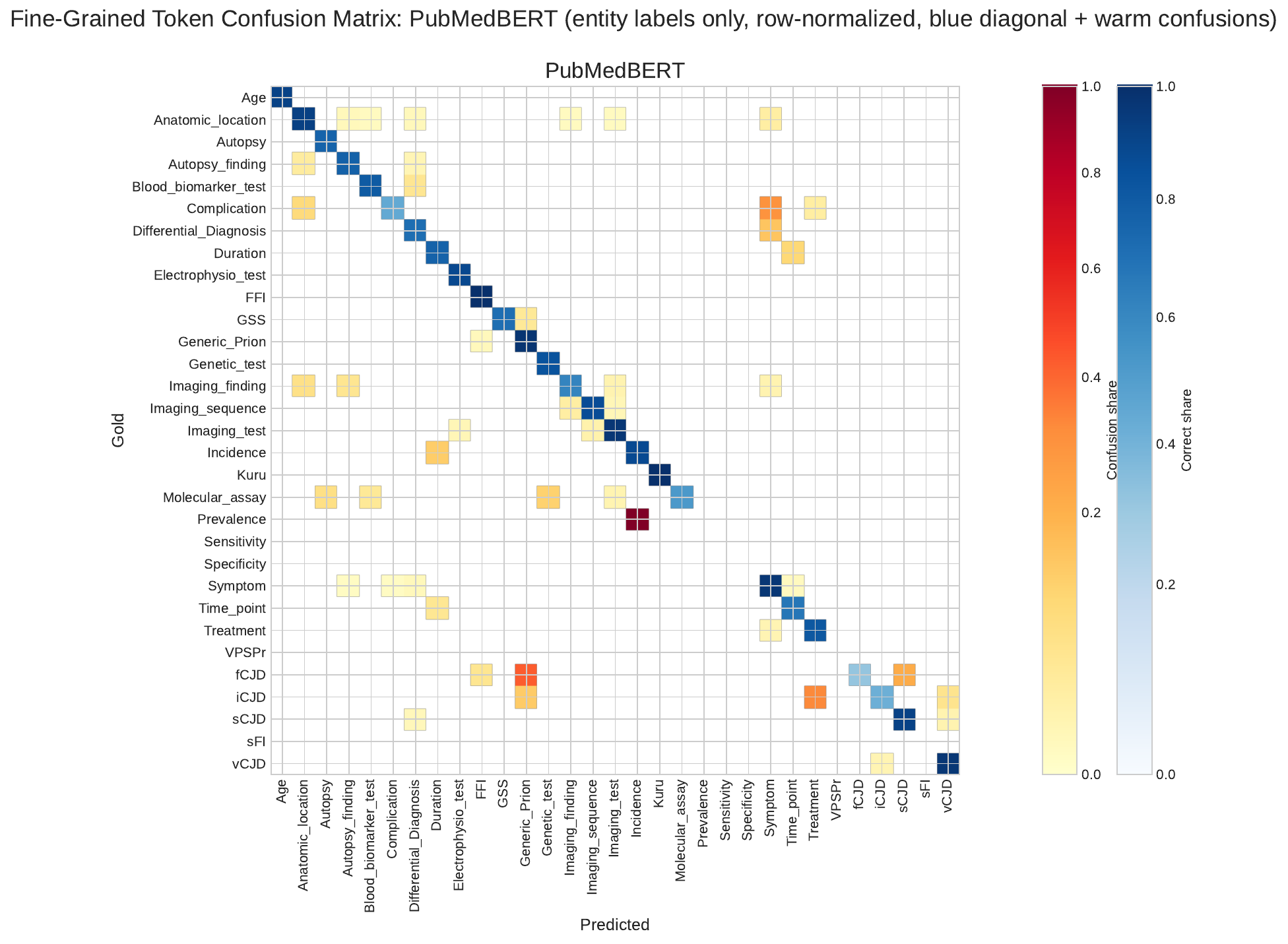}
\caption{PubMedBERT}
\label{fig:fine_pubmedbert_confusion_entity_only}
\end{subfigure}

\vspace{0.75em}

\begin{subfigure}[t]{0.82\textwidth}
\centering
\includegraphics[width=\linewidth]{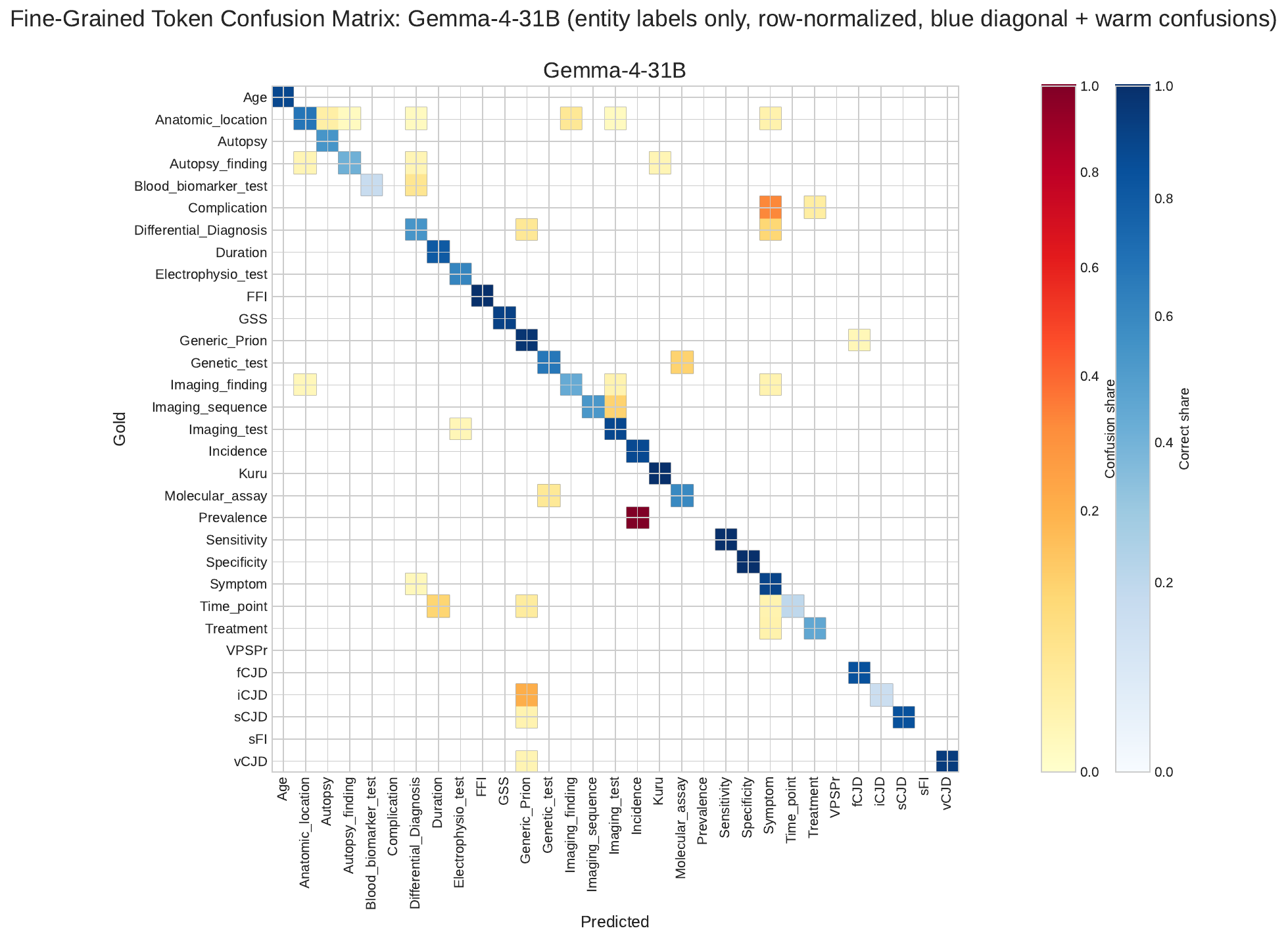}
\caption{Gemma-4-31B}
\label{fig:fine_gemma4_31b_confusion_entity_only}
\end{subfigure}
\caption{Entity-only fine-grained confusion matrices for the strongest supervised model (PubMedBERT) and the strongest zero-shot model (Gemma-4-31B).}
\label{fig:fine_entity_only_confusion_main_models}
\end{figure*}

Figures~\ref{fig:fine_pubmedbert_confusion_entity_only} and~\ref{fig:fine_gemma4_31b_confusion_entity_only} provide a complementary entity-only view of fine-grained errors for the two strongest models.
Under this evaluation, PubMedBERT is clearly stronger overall, reaching 0.794 flat entity-level F1 compared with 0.684 for Gemma-4-31B.
The main difference is recall: PubMedBERT reaches 0.793 recall versus 0.597 for Gemma-4-31B, while their precision remains very similar at 0.796 and 0.801, respectively.
This pattern shows that Gemma-4-31B is comparatively conservative, preserving Gemma-level precision but missing many more entities.

In these row-normalized, entity-label-only matrices, PubMedBERT has the cleaner diagonal overall.
More labels retain darker diagonal mass with less off-diagonal spill, indicating more consistent label assignment across the schema.
Gemma-4-31B looks reasonable on a few canonical labels, but its diagonal is visibly weaker for many harder clinical categories, including \textit{Autopsy}, \textit{Autopsy\_finding}, \textit{Blood\_biomarker\_test}, \textit{Time\_point}, and \textit{Treatment}.

Several confusion patterns are shared by both models.
\textit{Prevalence} and \textit{Incidence} remain difficult to separate, the imaging cluster shows substantial overlap among \textit{Imaging\_finding}, \textit{Imaging\_sequence}, and \textit{Imaging\_test}, and the prion subtype region also remains imperfectly separated, especially among \textit{fCJD}, \textit{iCJD}, \textit{GSS}, \textit{sFI}, and nearby subtype labels.
These are largely fine-grained semantic confusions between clinically adjacent categories rather than arbitrary label noise.
One recurring shared error is the prediction of \textit{Generic\_Prion} for \textit{fCJD} and \textit{iCJD}.
For \textit{fCJD}, a plausible explanation is that abstracts often realize the subtype as phrases such as ``familial CJD'' or ``familial form of CJD,'' where the diagnostically important modifier appears before the base disease name.
In these cases, the model appears to anchor on ``CJD'' while failing to consistently preserve the prefix that signals the familial subtype.
The same pattern likely explains part of the \textit{iCJD} $\rightarrow$ \textit{Generic\_Prion} confusion: the model often captures ``CJD'' but misses the preceding cue that indicates an iatrogenic or transmitted form.
This error is amplified by longer compositional mentions such as ``corneal transplant-related CJD,'' where the treatment-related phrase can be separated from the disease name; in such cases, the model may label ``corneal transplant'' as \textit{Treatment} and the remaining ``CJD'' span as \textit{Generic\_Prion} rather than assigning the full mention to \textit{iCJD}.
Another plausible shared confusion is between \textit{Complication} and \textit{Symptom}.
Some complication mentions are lexically symptom-like in isolation, but are labeled as \textit{Complication} because they denote downstream consequences of disease progression rather than primary manifestations.
For example, \textit{dysphagia} can be a symptom in many settings, but in prion disease abstracts it may appear as a late-stage secondary consequence of earlier neurological decline, which makes the boundary between the two labels difficult to recover from local context alone.
Similarly, mentions such as ``died'' or ``death'' may be pulled toward \textit{Symptom} or other outcome-like categories even though they are not direct symptomatic expressions of prion disease itself.
Instead, they typically reflect the final downstream result of severe functional deterioration and accumulated health complications over the course of the disease.

For PubMedBERT, the strongest diagonal cells are concentrated on common, high-volume labels such as \textit{Symptom}, \textit{Anatomic\_location}, \textit{Imaging\_test}, \textit{Generic\_Prion}, and \textit{sCJD}.
Its remaining errors mostly look like local mix-ups within semantically related label families rather than broad collapse across the label space.
Gemma-4-31B, by contrast, is weaker on context-heavy diagnostic labels and on imaging-related distinctions; its diagonal is often present but noticeably lighter, suggesting less stable fine-grained label assignment even when an entity span is recovered.

This entity-only view should also be read together with the full confusion analysis that includes \textit{O}.
Because these matrices exclude non-entity predictions, they likely understate Gemma-4-31B's main weakness, namely omission of the entity altogether.
In other words, PubMedBERT appears stronger and more balanced, with errors concentrated within semantically adjacent labels, whereas Gemma-4-31B shows reasonable zero-shot behavior but is less reliable on subtle distinctions and is likely more omission-heavy in the full setting.

At the label level, PubMedBERT remains strong on common clinical categories, including \textit{Generic\_Prion} (0.918), \textit{Imaging\_test} (0.908), \textit{sCJD} (0.897), \textit{vCJD} (0.887), \textit{Anatomic\_location} (0.883), and \textit{Symptom} (0.850).
Gemma-4-31B is still strong on a narrower set of distinctive labels, including \textit{Generic\_Prion} (0.917), \textit{FFI} (0.977), \textit{Kuru} (0.923), \textit{Age} (0.792), \textit{Imaging\_test} (0.781), and \textit{Symptom} (0.773), but its performance drops sharply on rarer or more context-dependent categories.
Its weakest flat F1 values include \textit{Complication} (0.000), \textit{Prevalence} (0.000), \textit{Sensitivity} (0.000), \textit{Specificity} (0.000), \textit{Blood\_biomarker\_test} (0.172), \textit{iCJD} (0.286), \textit{Time\_point} (0.317), and \textit{Imaging\_finding} (0.321).
PubMedBERT also shows weak spots, but they are milder and are concentrated in rare epidemiology labels and subtype distinctions: \textit{Prevalence} (0.000), \textit{Sensitivity} (0.000), \textit{Specificity} (0.000), \textit{fCJD} (0.333), \textit{Time\_point} (0.404), \textit{iCJD} (0.500), and \textit{Molecular\_assay} (0.528).
Overall, these figures reinforce the main result: PubMedBERT is the better model for PrionNER because it maintains Gemma-4-31B-level precision while recovering many more entities, whereas Gemma-4-31B remains the strongest zero-shot model but is still limited primarily by recall.

\subsection{Per-label Annotation Agreement}

Table~\ref{tab:per_label_agreement} provides the full per-label agreement breakdown for the test split.
Agreement is highest for common and semantically distinctive disease labels such as \textit{vCJD}, \textit{GSS}, \textit{FFI}, and \textit{sCJD}, while lower agreement is concentrated in sparse or boundary-sensitive categories such as \textit{Complication}, \textit{Prevalence}, \textit{Incidence}, and \textit{Imaging\_finding}.

\begin{table*}[t]
\centering
\small
\setlength{\tabcolsep}{3.5pt}
\resizebox{\textwidth}{!}{%
\begin{tabular}{lrrrrrr}
\toprule
\textbf{Label} & \textbf{F1} & \textbf{Matches} & \textbf{Ann.\ 1} & \textbf{Ann.\ 2} & \textbf{Union} & \textbf{Jaccard} \\
\midrule
Complication & 0.00 & 0 & 12 & 2 & 14 & 0.00 \\
Prevalence & 0.00 & 0 & 2 & 4 & 6 & 0.00 \\
Incidence & 40.00 & 1 & 2 & 3 & 4 & 25.00 \\
Imaging\_finding & 47.19 & 21 & 49 & 40 & 68 & 30.88 \\
Autopsy\_finding & 51.61 & 16 & 37 & 25 & 46 & 34.78 \\
Blood\_biomarker\_test & 52.63 & 20 & 32 & 44 & 56 & 35.71 \\
Time\_point & 60.61 & 20 & 34 & 32 & 46 & 43.48 \\
Treatment & 63.01 & 46 & 58 & 88 & 100 & 46.00 \\
Differential\_Diagnosis & 63.64 & 28 & 36 & 52 & 60 & 46.67 \\
Autopsy & 76.54 & 31 & 43 & 38 & 50 & 62.00 \\
Age & 76.92 & 20 & 26 & 26 & 32 & 62.50 \\
Duration & 77.65 & 33 & 40 & 45 & 52 & 63.46 \\
iCJD & 78.26 & 9 & 11 & 12 & 14 & 64.29 \\
Imaging\_sequence & 80.00 & 50 & 66 & 59 & 75 & 66.67 \\
Genetic\_test & 80.00 & 4 & 4 & 6 & 6 & 66.67 \\
Symptom & 81.00 & 194 & 235 & 244 & 285 & 68.07 \\
Specificity & 83.33 & 5 & 5 & 7 & 7 & 71.43 \\
Electrophysio\_test & 84.51 & 30 & 34 & 37 & 41 & 73.17 \\
Sensitivity & 85.71 & 6 & 6 & 8 & 8 & 75.00 \\
Molecular\_assay & 87.64 & 39 & 47 & 42 & 50 & 78.00 \\
Anatomic\_location & 90.30 & 149 & 159 & 171 & 181 & 82.32 \\
Imaging\_test & 92.31 & 60 & 64 & 66 & 70 & 85.71 \\
Kuru & 93.33 & 7 & 7 & 8 & 8 & 87.50 \\
Generic\_Prion & 94.83 & 220 & 234 & 230 & 244 & 90.16 \\
fCJD & 95.24 & 10 & 11 & 10 & 11 & 90.91 \\
sCJD & 95.54 & 75 & 79 & 78 & 82 & 91.46 \\
FFI & 95.65 & 22 & 22 & 24 & 24 & 91.67 \\
vCJD & 96.55 & 56 & 58 & 58 & 60 & 93.33 \\
GSS & 96.77 & 15 & 16 & 15 & 16 & 93.75 \\
\bottomrule
\end{tabular}
\vphantom{\big|}
}
\caption{Per-label annotation agreement under exact label-and-span matching on the 70-abstract test split, with one abstract excluded from the final agreement comparison.
Ann.\ 1 and Ann.\ 2 denote the number of entities annotated by Annotator 1 and Annotator 2, respectively, Matches denotes exact label-and-span matches, and Union is the size of the union of annotated entities for that label.}
\label{tab:per_label_agreement}
\end{table*}

\clearpage
\onecolumn
\section{Additional Reference Tables}

\subsection{Full Fine-grained Entity Distribution}

\begin{table}[t]
\centering
\small
\setlength{\tabcolsep}{6pt}
\begin{tabular}{llrrrr}
\toprule
\textbf{Coarse-grained Type} & \textbf{Fine-grained Type} & \textbf{Train Count} & \textbf{Train \%} & \textbf{Test Count} & \textbf{Test \%} \\
\midrule
Generic\_Prion & Generic\_Prion & 741 & 15.92 & 233 & 14.12 \\
Sporadic\_Prion & sCJD & 86 & 1.85 & 80 & 4.85 \\
Sporadic\_Prion & sFI & 3 & 0.06 & 0 & 0.00 \\
Sporadic\_Prion & VPSPr & 0 & 0.00 & 0 & 0.00 \\
Familial\_Prion & fCJD & 26 & 0.56 & 10 & 0.61 \\
Familial\_Prion & GSS & 32 & 0.69 & 16 & 0.97 \\
Familial\_Prion & FFI & 39 & 0.84 & 22 & 1.33 \\
Acquired\_Prion & vCJD & 102 & 2.19 & 55 & 3.33 \\
Acquired\_Prion & iCJD & 32 & 0.69 & 16 & 0.97 \\
Acquired\_Prion & Kuru & 28 & 0.60 & 7 & 0.42 \\
Differential\_Diagnosis & Differential\_Diagnosis & 188 & 4.04 & 36 & 2.18 \\
Symptom & Symptom & 791 & 16.99 & 248 & 15.03 \\
Test\_name & Imaging\_test & 261 & 5.61 & 75 & 4.55 \\
Test\_name & Electrophysio\_test & 173 & 3.72 & 38 & 2.30 \\
Test\_name & Blood\_biomarker\_test & 64 & 1.37 & 39 & 2.36 \\
Test\_name & Genetic\_test & 16 & 0.34 & 8 & 0.48 \\
Test\_name & Molecular\_assay & 23 & 0.49 & 45 & 2.73 \\
Test\_name & Autopsy & 199 & 4.27 & 64 & 3.88 \\
Sequences & Imaging\_sequence & 117 & 2.51 & 77 & 4.67 \\
Findings & Imaging\_finding & 153 & 3.29 & 71 & 4.30 \\
Findings & Autopsy\_finding & 215 & 4.62 & 42 & 2.55 \\
Anatomic\_location & Anatomic\_location & 646 & 13.88 & 223 & 13.52 \\
Age & Age & 137 & 2.94 & 29 & 1.76 \\
Treatment & Treatment & 126 & 2.71 & 88 & 5.33 \\
Complication & Complication & 78 & 1.68 & 19 & 1.15 \\
Time & Duration & 217 & 4.66 & 49 & 2.97 \\
Time & Time\_point & 152 & 3.27 & 47 & 2.85 \\
Stats & Sensitivity & 0 & 0.00 & 4 & 0.24 \\
Stats & Specificity & 0 & 0.00 & 4 & 0.24 \\
Stats & Prevalence & 4 & 0.09 & 1 & 0.06 \\
Stats & Incidence & 6 & 0.13 & 4 & 0.24 \\
\midrule
\textbf{Total} & & \textbf{4655} & \textbf{100.00} & \textbf{1650} & \textbf{100.00} \\
\bottomrule
\end{tabular}
\caption{Full fine-grained entity distribution in the PrionNER train and test splits.
Percentages are computed separately within the train and test sets over schema-defined fine-grained entity mentions only.}
\label{tab:checked_finegrained_distribution_train_test}
\end{table}

\subsection{Top Surface Forms by Entity Type}

Table~\ref{tab:entity-surface-dictionary-top-10} reports the 10 most frequent normalized surface forms for each entity type.
Numbers in parentheses after each surface form indicate mention counts, and the left-column notation ``U./M.'' denotes total unique surface forms and total mentions for that entity type.

\footnotesize
\setlength{\LTleft}{0pt}
\setlength{\LTright}{0pt}
\setlength{\LTcapwidth}{\textwidth}
\setlength{\emergencystretch}{2em}
\begin{center}
\captionsetup{hypcap=false}
\captionof{table}{Top 10 normalized surface forms for each entity type in the combined dataset.}
\label{tab:entity-surface-dictionary-top-10}
\end{center}
\begin{longtable}{@{}>{\raggedright\arraybackslash}p{0.19\linewidth}>{\raggedright\arraybackslash}p{0.75\linewidth}@{}}
\toprule
Entity Type (U./M.) & Top Surface Forms (mentions) \\
\midrule
\endfirsthead
\toprule
Entity Type (U./M.) & Top Surface Forms (mentions) \\
\midrule
\endhead
\midrule
\multicolumn{2}{r}{Continued on next page} \\
\midrule
\endfoot
\bottomrule
\endlastfoot
Symptom\\(524 / 1279) & dementia (79); myoclonus (79); ataxia (47); rapidly progressive dementia (36); psychiatric symptoms (24); akinetic mutism (16); progressive dementia (16); myoclonic jerks (15); aggression (15); neurological symptoms (14) \\
Anatomic\_location\\(276 / 837) & brain (48); basal ganglia (41); white matter (29); cortical (26); cerebellum (23); cerebellar (23); thalamus (22); striatum (21); cerebral cortex (19); pulvinar (18) \\
Duration\\(201 / 290) & 7 months (8); within a year (7); 12 months (7); four months (6); 13 months (6); within one year (5); 4 months (5); 3 months (5); 5 months (5); disease duration (4) \\
Time\_point\\(181 / 248) & age at onset (5); at onset (5); 1996 (5); after onset (5); in 1996 (5); 1985 (4); at the onset (4); the end of the third stage (4); clinical onset (3); early in the course (3) \\
Imaging\_finding\\(161 / 237) & pulvinar sign (17); brain atrophy (13); hyperintensity (11); high signal intensity (7); atrophy (6); hyperintense signal abnormalities (4); signal intensity abnormalities (4); high signal (4); hyperintensities (3); cortical atrophy (3) \\
Autopsy\_finding\\(155 / 276) & neuronal loss (22); spongiform change (20); spongiosis (9); spongiform changes (8); gliosis (8); status spongiosus (8); astrocytosis (7); neuronal degeneration (5); spongiform encephalopathy (5); kuru plaques (5) \\
Differential\_Diagnosis\\(136 / 278) & alzheimer's disease (17); dlb (16); dementia (14); ad (12); stroke (9); neurodegenerative disorder (7); non‐prion disorders (6); insomnia (5); dementias (4); thalamic dementia (4) \\
Treatment\\(131 / 241) & palliative care (23); quinacrine (16); antipsychotics (9); quetiapine (6); haloperidol (6); treatment (4); anesthesia (4); supportive care (4); hospice care (4); dura mater graft (4) \\
Age\\(124 / 182) & 61-year-old (5); elderly (5); 59-year-old (5); 49-year-old (5); 54-year-old (4); 65‐year old (3); 48‐year‐old (3); 70-year-old (3); 75-year-old (3); 58-year-old (3) \\
Imaging\_test\\(100 / 383) & mri (82); magnetic resonance imaging (37); ct (28); spect (27); mr imaging (17); mr (13); mr images (13); ct scan (9); positron emission tomography (7); brain mri (7) \\
Generic\_Prion\\(85 / 1219) & cjd (576); creutzfeldt-jakob disease (327); prion disease (50); bse (33); prion diseases (30); bovine spongiform encephalopathy (24); spongiform encephalopathy (16); creutzfeldt‐jakob disease (14); creutzfeldt-jacob disease (9); subacute spongiform encephalopathy (6) \\
Autopsy\\(74 / 271) & autopsy (62); brain biopsy (36); necropsy (20); biopsy (13); neuropathological (11); neuropathological examination (9); neuropathologically (7); pathological (5); pathologically (5); histologically (5) \\
Imaging\_sequence\\(63 / 241) & dwi (34); flair (34); diffusion-weighted (15); diffusion-weighted imaging (12); t2-weighted (12); t2 (11); diffusion-weighted images (9); dw (8); adcs (8); dw images (8) \\
Electrophysio\_test\\(49 / 247) & eeg (119); electroencephalogram (29); electroencephalographic (15); electroencephalography (10); eegs (8); electroencephalograms (6); erg (4); polysomnography (4); br (3); electroencephalographic findings (3) \\
Blood\_biomarker\_test\\(44 / 140) & csf (57); cerebrospinal fluid (14); blood tests (9); csf analysis (6); csf tau protein (4); csf virology (3); biomarkers (3); csf tau-pt181 (2); csf studies (2); serum workup (2) \\
Molecular\_assay\\(40 / 111) & cdi (16); csf rt‐quic (6); immunohistochemistry (6); ihc (6); western blot (5); immunocytochemistry (5); molecular analysis (4); densitometric analysis (4); protein assay (3); csf rt‐qulc (3) \\
sCJD\\(26 / 241) & scjd (90); sporadic cjd (52); sporadic creutzfeldt-jakob disease (29); sporadic (17); vv2 (8); heidenhain variant (4); creutzfeldt-jakob disease (4); sporadic form (4); mv1 (4); mm1 (4) \\
Complication\\(22 / 52) & death (20); died (7); myocarditis (3); deaths (2); bronchopneumonia (2); fatal (2); dysphasia (1); loss of independence (1); dead (1); acute myocarditis (1) \\
iCJD\\(19 / 51) & iatrogenic cjd (8); iatrogenic creutzfeldt-jakob disease (8); dcjd (8); icjd (5); iatrogenic (4); iatrogenic forms (3); iatrogenic cases (2); dural graft associated cjd (2); iatrogenic transmission (1); iatrogenic transmission of cjd (1) \\
GSS\\(18 / 63) & gss (19); gerstmann-sträussler-scheinker disease (14); gss102 (6); gerstmann-straussler-scheinker disease (4); gss105 (4); gssd (3); gerstmann-straussler-scheinker syndrome (2); gerstmann-sträussler-scheinker's disease (1); gerstmann-sträussler-scheinker's syndrome (1); gerstmann-strässler-scheinker's syndrome (1) \\
vCJD\\(18 / 218) & vcjd (104); nvcjd (37); variant creutzfeldt-jakob disease (26); variant cjd (26); nv-cjd (4); acquired (4); variant creutzfeldt-jakob (2); infectious (2); new variant cjd (2); variant (2) \\
fCJD\\(16 / 47) & familial (9); familial cjd (8); inherited prion disease (6); inherited (4); fcjd (3); genetic cjd (3); genetic (3); hereditary cjd (2); cjd178 (2); genetic forms (1) \\
Genetic\_test\\(15 / 26) & molecular genetic analysis (5); genetic testing (3); genetic analysis (2); prp gene analysis (2); restriction-enzyme analysis (2); analysing dna (2); prion gene analysis (2); genetic examination (1); genotyping (1); genetic tests (1) \\
Incidence\\(11 / 12) & high incidence (2); 0.06\% in 2023 (1); 0.10\% in 2024 (1); 1 to 2 cases per million people per year (1); annual incidence of 0.37 cases/million (1); incidence (1); annual incidence of 0.5-1.5 cases of cjd per million (1); incidence of approximately .5-1 new cases per million population per year (1); incidence 1 in 1 000 000 (1); 1 to 2 cases per million people per year. (1) \\
Prevalence\\(9 / 10) & 1–2 people per million annually (2); 1–2/million/year (1); one case per million people per year (1); one in one million (1); 1 to 2 cases per million people per year (1); 35\% (1); 1–2 people per million annually. (1); 0.06\% (1); 0.10\% (1) \\
FFI\\(7 / 85) & ffi (44); fatal familial insomnia (33); ffi-1 (4); ffi-2 (1); fatal familiar insomnia (1); met-met subtype (1); fatal insomnia (1) \\
Sensitivity\\(7 / 15) & sensitivity (4); sensitivity of 100\% (2); sensitivity of 87\% (2); sensitivity (96\%) (2); 91\% sensitive (2); sensitivity higher (2); sensitive (1) \\
Specificity\\(5 / 12) & specificity (4); specificity of 92\% (2); specificity of 97\% (2); specificity (97\%) (2); 95\% specific (2) \\
Kuru\\(3 / 43) & kuru (40); kuru plaques (2); kuru type (1) \\
sFI\\(3 / 3) & ffi-1 (1); ffi-2 (1); fatal familial insomnia (1) \\
\end{longtable}

\end{document}